\pdfoutput=1
\documentclass[11pt]{article}
\usepackage{graphicx}

\usepackage[]{acl}

\usepackage{times}
\usepackage{latexsym}
\usepackage{natbib}
\usepackage{algorithm}
\usepackage[T1]{fontenc}
\usepackage[utf8]{inputenc}

\usepackage{microtype}
\usepackage{caption,subcaption}
\usepackage{graphicx}
\usepackage{amsmath}
\usepackage{booktabs}
\usepackage{multirow, epstopdf}
\usepackage{lineno}
\usepackage{amssymb}
\usepackage{placeins}
\usepackage{array}
\usepackage{longtable}
\usepackage{multirow}
\usepackage{pgfplots}
\usepackage{pgfplotstable}
\pgfplotsset{compat=1.18}

\title{\textit{Overcoming Black-box Attack Inefficiency with Hybrid and Dynamic Select Algorithms}}

\author {Abhinay Shankar Belde,  Rohit Ramkumar, Jonathan Rusert \\Department of Computer Science\\ Purdue University, Fort Wayne\\ 
\texttt{\{belda01, ramkr01, jrusert\}@pfw.edu}}
\usepackage{hyperref}
\usepackage{algorithm}
\usepackage{algorithmic}
\usepackage{amsmath}
\begin{document} 
\maketitle

\vspace{1cm}

\begin{abstract}
Adversarial text attack research plays a crucial role in evaluating the robustness of NLP models. However, the increasing complexity of transformer-based architectures has dramatically raised the computational cost of attack testing, especially for researchers with limited resources (e.g., GPUs). Existing popular black-box attack methods often require a large number of queries, which can make them inefficient and impractical for researchers. To address these challenges, we propose two new attack selection strategies called Hybrid and Dynamic Select, which better combine the strengths of previous selection algorithms. Hybrid Select merges generalized BinarySelect techniques with GreedySelect by introducing a size threshold to decide which selection algorithm to use. Dynamic Select provides an alternative approach of combining the generalized Binary and GreedySelect by learning which lengths of texts each selection method should be applied to.   This greatly reduces the number of queries needed while maintaining attack effectiveness (a limitation of BinarySelect). Across 4 datasets and 6 target models,
our best method(sentence-level Hybrid Select) is able to reduce the number of required queries per attack up 25.82\% on average against both encoder models and LLMs, without losing the effectiveness of the attack. 
\end{abstract}
% When evaluated against five classifiers across three datasets, our method consistently reduces the number of queries—yielding a 28 percent decrease in the number of queries compared to traditional greedy search on the IMDb dataset, with only a modest 4-point decrease in attack success rate. This approach paves the way for more accessible adversarial research by making attack testing feasible for researchers with fewer computational resources, promoting inclusivity and innovation in the study of NLP robustness.

\section{Introduction}
\begin{figure*}
    \centering
    \includegraphics[width=\textwidth]{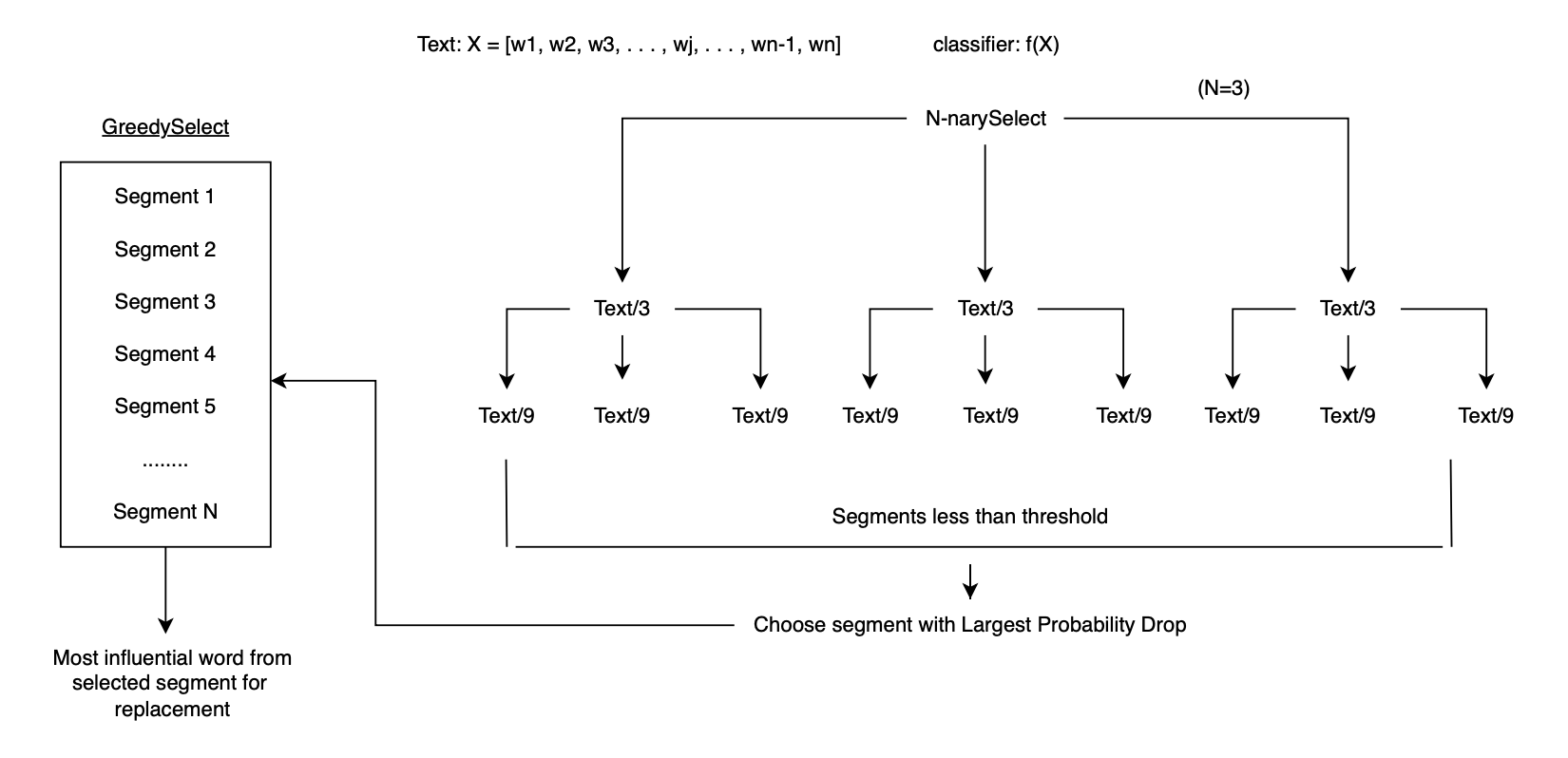}
    \caption{Visualization of HybridSelect. This figure illustrates the HybridSelect algorithm, which combines GreedySelect and N-narySelect techniques for efficient query optimization. The process begins with segmenting the input text \(X\) into smaller components. The GreedySelect method evaluates individual segments iteratively to determine their influence, identifying the most impactful word for replacement. Concurrently, N-narySelect partitions the text hierarchically, dividing it into \(n\)-segments (e.g., \(N=3\)), and evaluates subqueries using a classifier \(f(X)\). Probabilities are calculated, and the segment with the largest probability drop is selected. Segments below the threshold are finalized, ensuring an optimal balance between exploration and computational efficiency.} 
    \label{fig:hybridnary}
\end{figure*}

Adversarial attacks have been useful for studying the robustness of NLP models. Black-box world-level attacks often follow a two-step approach, first selecting a word or token to modify and then modifying that word to cause the target NLP model to fail (with further constraints). Selection algorithms can play a pivotal role in the computational overhead of adversarial attacks. GreedySelect \cite{hsieh-etal-2019-robustness, li-etal-2020-bertattack}, a widely adopted method, iteratively evaluates each word in the text to determine its contribution to the classifier's decision. By removing/masking the word with the highest probability drop, GreedySelect achieves high precision. However, its word-by-word approach often results in high query counts, limiting its efficiency. 

As NLP models continue to grow in size, high query counts can translate to attacks taking a large of amount of time per text. This inefficiency further affects researchers or users who do not have access to high-end GPU's to carry out attacks. Furthermore, an increased time in attack execution, reduces the ability for researchers to adequately test their models.

To address these challenges, BinarySelect \cite{ghosh-rusert-2025-binaryselect} adopted a divide-and-conquer strategy inspired by binary search. By partitioning the input into two segments and evaluating the probability change caused by excluding each segment, BinarySelect aimed to narrow down the search space when selecting an influential word. While BinarySelect helped reduced the number of queries compared to GreedySelect in some instances, it caused a drop in attack effectiveness.

Building on these foundations, this paper proposes hybrid approaches that integrate query-efficient methods like BinarySelect with GreedySelect's effectiveness. These strategies aim to balance attack efficacy and computational efficiency, addressing critical gaps identified in prior literature and offering a solution to improve adversarial attack performance.

Our research makes the following contributions:

\begin{enumerate}
    % \item \textbf{Propose Two New Selection Algorithm, Hybrid and Dynamic Select:} 
    % We introduce two main algorithms which combines a generalized BinarySelect with GreedySelect. We call the generalized BinarySelect ``N-nary'' for simplicity. 
    % \textbf{Hybrid Select} leverages N-nary up to a threshold and then switches to GreedySelect. \textbf{Dynamic Select} learns for which lengths of text which $N$'s are most effective and efficient, and uses this knowledge to its advantage. Additionally, we combine both methods to leverage the strengths of both. Overall, we find improvements over BinarySelect and GreedySelect with various combinations of all three methods. For example, one variation of \textbf{Hybrid Select} reduces the average query count up to \textbf{16.7\%} (e.g., from 433 to 371 queries), while also maintaining attack effectiveness.
    \item \textbf{New Selection Algorithms (Hybrid and Dynamic Select).} 
    We propose two new algorithms that extend BinarySelect and GreedySelect. \textit{Hybrid Select} uses N-nary partitioning up to a threshold before switching to GreedySelect, while \textit{Sentence-Level Hybrid} is a special case that partitions by sentence. \textit{Dynamic Select} instead learns which N values are most effective for different text lengths, adapting the strategy automatically. We also evaluate their combination (\textit{Hybrid + Dynamic}), highlighting where it helps and where it fails. Across 4 datasets and 6 models, one variant of \textit{Sentence-Level Hybrid} reduces query counts by up 25.82\% (e.g., 484 $\rightarrow$ 362 queries~(Table~\ref{tab:imdb_results})) while preserving attack success. Unlike prior works that focus on improving the \textbf{replacement} stage, our contribution advances the \textbf{selection} step and generalizes to many greedy-based attacks (e.g., TextFooler, BERT-Attack, PWWS)\footnote{We share our code for reproducibility: \url{https://github.com/08Abhinay/Hybrid-Dynamic-Select}}.
   
    \item \textbf{Evaluate Methods Across Multiple Datasets and Dimensions:} 
     We evaluate our methods on adversarial attacks against both encoder models (DistilBert and DeBERTa) and LLMs (GPT2, GPTNeo, T5, and Llama3) across four standard adversarial attacks classification datasets of various sizes: IMDB, Yelp, AG News, and Rotten Tomatoes. Additionally, we conduct several ablation studies to find and demonstrate the strongest hyperparameters and confirm the effectiveness and efficiency of our methods.

\end{enumerate}

Overall, our proposed methods improve efficiency over previous methods like GreedySelect while maintaining efficacy, which was not achieved by previous methods such as BinarySelect.

\section{Proposed Approaches}
Our proposed approaches focus on the domain of black-box attacks (sometimes noted as grey-box). That is, following previous research \cite{li2021contextualized,  hsieh-etal-2019-robustness, alzantot2018generating, gao2018blackbox}, we assume information about the model, such as weights and architecture, are not known. The attack can pass a query to the model, which returns a prediction with a probability. 

The goal of the attacker is to modify an input text, such that it causes a target model to fail, but retains its original meaning to humans \cite{li-etal-2020-bert-attack, garg2020bae}. In this research, we focus on the domain of text classification. Often black-box word-level text attacks consist of 2 stages: selection and modification. In selection,
the attacker finds the word or token that the classifier is relying on the most for its decision. In modification, the token found in the selection step is modified in such a way to keep the noted goals. This can be character level modification or word level by replacing the token with less known synonyms.  If the attack does not succeed after the modification, the selection method finds the next influential word and the process continues.

In this research, our objective is to improve the efficiency of attacks via the selection step by balancing the efficiency of BinarySelect and the attack effectiveness of GreedySelect. We first define GreedySelect and BinarySelect and then our own approaches.

%In particular, we focus on creating efficient selection methods.
%explored are N-Nary Select, Greedy Select, Hybrid method combining N-Nary and Greedy and Dynamic N. The replacement method explored are BERT-on-BERT replacement method (TODO: Add reference).

%\input{threatmodel}

%Selection method is used to find the replacement word which would be able to flip the results of the classifier. The replacement word is the most influential word in the input text, which is the word that has more impact in the prediction of the classifier.

\subsection{Baseline Selection Algorithms}

\paragraph{Baseline selection.}
\emph{Greedy Select} (and its variations such as Importance Score) is one of the most widely used method for black-box text attacks \cite{formento-etal-2023-using, jin2020bert,li-etal-2020-bert-attack, garg2020bae, ren-etal-2019-generating, hsieh-etal-2019-robustness, Li2019Textbugger, gao2018blackbox}. Greedy Select removes one word at a time, queries the classifier, and keeps the word whose omission produces the largest drop in the target‑class probability.  
Because it probes every token, it is highly effective but also query‑intensive (one query per word just to find the first salient token).  
\emph{Binary Select}~\cite{ghosh-rusert-2025-binaryselect} eases this burden by applying a binary‑search heuristic: the sentence is repeatedly split in half, and the half whose removal hurts the classifier most is subdivided until only a single word remains.  
This strategy cuts the query cost dramatically on long texts but—by discarding half of the context at each step—can overlook subtle but important words, leaving a small effectiveness gap relative to Greedy Select.

\paragraph{}
To bridge that gap we introduce \emph{N‑nary Select}, which generalises Binary Select from two to \(N\) segments.  
Starting with the full input \(X\) and its score \(f(X)\), we partition \(X\) into \(N\) disjoint segments \(\{X_1,\dots,X_N\}\).  
For each segment we compute \(\Delta_i = f(X)-f(X\!\setminus\!X_i)\) and recurse on the segment with the largest \(\Delta_i\).  
The recursion stops when a single‑word segment is reached; that word is deemed most influential.  
By tuning \(N\), the algorithm interpolates smoothly between the thoroughness of Greedy Select (\(N=|X|\)) and the speed of Binary Select (\(N=2\)), allowing us to find a sweet‑spot that is both query‑efficient and consistently effective across text lengths and model architectures.

\subsection{Proposed Selection Methods}\label{sect:nnary_word}
Though \textit{N-nary Select} is more flexible than BinarySelect, our experiments show, it still suffers from issues found in BinarySelect. Namely, on small texts, the model is better off leveraging GreedySelect as the division of texts can cause extra unnecessary queries. Thus, we propose \textit{Hybrid Select} and \textit{Dynamic Select} to address this.

\subsubsection{\textit{Hybrid  Select}}\label{sect:nns_hybrid}
\textit{Hybrid  Select} combines \textit{N-nary Select} with GreedySelect by setting a length threshold $t$. The full algorithm can be found in Algorithms~\ref{alg:dynN}--\ref{alg:hybrid-pure}. Specifically, \textit{Hybrid  Select} first follows \textit{N-nary Select} and continuously partitions the texts into $N$ partitions. Once the length of the current partition falls below $t$, the algorithm transitions to word-level greedy exploration to pinpoint the most impactful words. The threshold $t$ is set as a percentage of the original text length. For example, if the original text is 200 words in length and $t$ is set to 10\%, then once the partitioned texts contain 20 words or less, the algorithm switches to GreedySelect. This combination of N-nary segmentation and word-level precision balances computational efficiency with attack effectiveness. A visualization of this method can be found in Figure \ref{fig:hybridnary}.

\noindent \textbf{\textit{Sentence Level Hybrid Select}:}
Another issue with  the \textit{N-nary Select} algorithm is that splitting the text in N separate segments can cause sentences to split and meaning affected. Thus, we also propose a special case of the \textit{Hybrid N-nary Select} methodology which splits the original text at the sentence level~(Alg~\ref{alg:hyb-sent}). This variation sets $N = S$, where $S$ is the number of sentences in the original text. The method then follows the standard Hybrid procedure: if any sentence segment remains longer than the threshold $t$, it continues N-nary splitting; once the length drops below $t$, it switches to GreedySelect. In practice, this means the method first identifies the most influential sentence via \textit{N-nary Select}, and then pinpoints the most influential words within that sentence via GreedySelect.
% This variation essentially sets $N$ to the number of sentences in the original text, by splitting the entire text at the sentence level first.
% The method then follows the normal hybrid of splitting the sentence by $N$ until it hits the thresholds $t$ at which point it switches to GreedySelect.  Thus, this method first explores the most influential sentence via \textit{N-nary Select} and then finds the most influential word via GreedySelect. 

The hybrid nature of this approach enables it to leverage coarse-grained sentence segmentation for rapid narrowing, followed by fine-grained word exploration for precision. It is particularly effective in scenarios where sentence boundaries provide a natural structure to the input text.

\subsubsection{\textit{Dynamic Select}}
We also propose an algorithm which leverages GreedySelect and \textit{N-nary Select} on the texts they are the most effective on. That is, we find the optimal N value based on the length of the input text in words and then use this in the attack. As shown in Appendix \ref{DynamicN_bins}, these values are calculated from a separate validation set in the same dataset. Initially, the maximum input length in the validation set is identified. Based on this length, 5 length bins are created. For each length bins, the optimal N value is selected based on the minimum query count in the validation set. This mapping of length bins to the optimal N value is applied to the target data.

\subsubsection{\textit{Hybrid + Dynamic Select}}
Finally, we aim to combine the strengths of both \textit{Hybrid Select} and \textit{Dynamic Select} into one methodology \textit{Hybrid + Dynamic Select}. This method combines both algorithms by first using \textit{Dynamic Select} to find the optimal $N$ and then passing that $N$ to the \textit{Hybrid Select} methodology. Specifically, the algorithm will automatically choose the $N$ value based on the binning rules shown in Algorithm~\ref{alg:dynN}.

%{\color{blue} We also give the user the option to set n\_mode = auto in the CLI commands, which will automatically choose the $N$ value based on the binning rules shown in Algorithm~\ref{alg:dynN}. Alternatively, the user may set $N$ to any arbitrary value of their choice.}

\subsection{Replacement Methodology}
As our proposed improvements focus on the selection algorithm, we leverage a simple replacement algorithm leveraged by previous research \cite{ren-etal-2019-generating}. We leverage Wordnet \cite{miller-1994-wordnet} to obtain synonyms of selected words and try to replace the word and check the change in probability. We begin with the most similar synonym (as given by Wordnet) and continue checking every synonym until either the classifier fails or all have been checked. If the classifier did not fail, we choose the synonym that causes the highest drop in target label probability. We also perform experiments to see if utilizing a transformer model for masked language modeling replacement would improve the attack effectiveness or efficiency (similar to previous research \cite{li-etal-2020-bertattack, garg2020bae}). These can be found in Section \ref{sect:wordreplacement}.

\begin{table*}[!h]
\resizebox{\textwidth}{!}{%
\renewcommand{\arraystretch}{1.5}
\begin{tabular}{|c|c|c|c|c|c||c|c|c|c|c|c|c|c|}
\hline

 & & \multicolumn{2}{|c|}{\textbf{DistilBERT} (OA = 92)} & \multicolumn{2}{|c|}{\textbf{DeBERTa} (OA = 94)} & \multicolumn{2}{|c|}{\textbf{GPT-2} (OA = 94)} & \multicolumn{2}{|c|}{\textbf{GPT-Neo} (OA = 93)} & \multicolumn{2}{|c|}{\textbf{T5} (OA = 90)} & \multicolumn{2}{|c|}{\textbf{Llama3} (OA = 92)} \\\hline
\textbf{Selection Method} & \textbf{$N$} & \textbf{ASR} & \textbf{AvgQ} & \textbf{ASR} & \textbf{AvgQ} & \textbf{ASR} & \textbf{AvgQ} & \textbf{ASR} & \textbf{AvgQ} & \textbf{ASR} & \textbf{AvgQ} & \textbf{ASR} & \textbf{AvgQ} \\ \hline

Binary & - & \underline{98} & 484 & 71 & 372 & 90 & 652 & 98 & 583 & 98 & 480 & 98 & 547 \\ \cline{1-2} \cline{3-14}

Greedy & - & \textbf{99} & 425 & 70 & \underline{303} & 90 & 593 & 98 & 462 & 99 & 412 & 98 & 430 \\ \cline{1-2} \cline{3-14} \hline

\multirow{2}{*}{N-Nary} & 3 & \underline{98} & 461 & 70 & 372 & 93 & 614 & 98 & 537 & 99 & 455 & 100 & 492 \\ \cline{2-2} \cline{3-14}

& 6 & \underline{98} & 569 & 71 & 447 & 92 & 831 & 99 & 643 & 99 & 496 & 98 & 581 \\ \cline{1-2} \cline{3-14}

\multirow{3}{*}{Hybrid} & 2 & \underline{98} & 427 & 70 & 333 & 99 & 507 & 98 & 485 & 99 & \underline{391} & 98 & 458 \\ \cline{2-2} \cline{3-14}

& 3 & \underline{98} & \underline{392} & 70 & \underline{302} & 98 & 475 & 97 & \underline{457} & 99 & \underline{392} & 100 & \underline{427} \\ \cline{2-2} \cline{3-14}

& 6 & \underline{98} & 398 & 70 & 360 & 99 & 533 & 99 & 555 & 99 & 422 & 98 & 504 \\ \cline{1-2} \cline{3-14}

\multirow{3}{*}{Sent. Hybrid} & 2 & \underline{98} & \underline{368} & 71 & 303 & 98 & 430 & 97 & 456 & 98 & 384 & 97 & 459 \\ \cline{2-2} \cline{3-14}

& 3 & \underline{98} & \textbf{362} & 71 & \underline{302} & 98 & 443 & 98 & 445 & 98 & 368 & 97 & 454 \\ \cline{2-2} \cline{3-14}

& 6 & \underline{98} & 373 & \textbf{71} & \textbf{295} & 98 & \textbf{428} & 97 & \textbf{430} & 98 & \textbf{359} & 97 & 436 \\ \cline{1-2} \cline{3-14}

Dynamic & - & \underline{98} & 456 & 70 & 384 & 98 & 552 & 98 & 553 & 99 & 429 & 98 & 497 \\ \cline{1-2} \cline{3-14}

Dynamic + Hybrid & - & \textbf{99} & 395 & 71 & 312 & 69 & 463 & 73 & 467 & 99 & 376 & 98 & \textbf{415} \\ \hline

\textbf{Selection Method} & \textbf{$N$} &\textbf{Sim.} & \textbf{Pert.} &\textbf{Sim.} & \textbf{Pert.} & \textbf{Sim.} & \textbf{Pert.} & \textbf{Sim.} & \textbf{Pert.} &\textbf{Sim.} & \textbf{Pert.} & \textbf{Sim.} & \textbf{Pert.} \\ \hline

Binary  & - & 91 & 9 & 92 & 8  & 84 & 13 & 90 & 12 & 89 & 11 & 91 & 10 \\ \cline{1-2} \cline{3-14}

Greedy  & - & 91  & 7 & 92 & 7  & 85 & 12 & 91  & 8 & 90  & 8 & 92 & 7 \\ \cline{1-2} \cline{3-14} 

\multirow{2}{*}{N-Nary } & 3  & 91 & 8 & 91 & 8  & 84 & 14 & 90 & 11 & 89 & 11 & 91 & 10 \\ \cline{2-2} \cline{3-14}

& 6   & 91  & 7  & 92  & 7 & 85 & 13 & 91 & 9 & 90 & 9 & 92 & 8\\ \cline{1-2} \cline{3-14}

\multirow{3}{*}{Hybrid} & 2 & 92  & 9 & 92 & 8 & 89 & 12 & 90 & 11 & 89 & 11 &  91 & 10 \\ \cline{2-2} \cline{3-14}

& 3 & 92  &8 &  91 & 8 &  89 & 11  & 90 & 11 & 89 & 11 & 91 & 10 \\ \cline{2-2} \cline{3-14}

& 6  & 92  & 7  & 92 & 7 & 90 & 9 & 91 & 9 & 90 & 9 & 92 & 8\\ \cline{1-2} \cline{3-14}

\multirow{3}{*}{Sent. Hybrid} & 2   & 91  & 9 &  91 & 9  & 89 & 11 & 90 & 11 & 89 & 12 & 90 & 11 \\ \cline{2-2} \cline{3-14}

& 3  & 91  & 9  & 90 & 8  & 89 & 11 & 90 & 11 & 89 & 11 & 90 & 10\\ \cline{2-2} \cline{3-14}

& 6 & 91  & 9 & 91 & 7 & 90 & 10 & 91 & 10 & 89 & 11 & 90 & 10\\ \cline{1-2} \cline{3-14}

Dynamic & - & 91  & 8 & 82  & 19  &89  &11  &90 &11 &90 & 9& 91&9 \\ \cline{1-2} \cline{3-14}

Dynamic + Hybrid & -  & 91  & 8  & 82  & 19  & 89&11 & 90& 11&90 &9 & 91 & 9\\ 
\hline

\end{tabular}%
}
\caption{Results for different methods on the IMDB dataset where ASR is attack success rate, AvgQ is the average queries by the successful attacks, Sim. is the similarity of text to the original, and Pert. is the percentage of text modified. A $t$ of 10\% was used for all Hybrid methods. \textbf{Bold} values indicate the best for that column, \underline{underline} indicates second best.}
\label{tab:imdb_results}
\end{table*}

\begin{table*}[!h]
\resizebox{\textwidth}{!}{%
\renewcommand{\arraystretch}{1.5}
\begin{tabular}{|c|c|c|c|c|c||c|c|c|c|c|c|}
\hline
 & & \multicolumn{2}{|c|}{\textbf{DistilBert} (OA = 97)} & \multicolumn{2}{|c|}{\textbf{DeBERTa} (OA = 99)} & \multicolumn{2}{|c|}{\textbf{GPT2} (OA = 96)} & \multicolumn{2}{|c|}{\textbf{GPTNeo} (OA = 97)} & \multicolumn{2}{|c|}{\textbf{T5} (OA = 97)}  \\\hline
\textbf{Selection Method} & \textbf{$N$} &\textbf{ASR} & \textbf{AvgQ} & \textbf{ASR} & \textbf{AvgQ} & \textbf{ASR} & \textbf{AvgQ} & \textbf{ASR} & \textbf{AvgQ} & \textbf{ASR} & \textbf{AvgQ} \\ \hline
Binary  & - &   \underline{97} & 400 & 98 & 467 & 97 & 461 & 96 & 368 & 69 & 602 \\ \cline{1-2} \cline{3-12}

Greedy  & -   &  \textbf{98} & 329 &  95 &  \underline{462} & 96 & \textbf{355} & 97 & 323 & 74 & \textbf{525}\\ \cline{1-2} 
\cline{3-12} \hline

\multirow{2}{*}{N-Nary } & 3 & \underline{97} & 374 & 95 & 469 & 96 & 426 & 96 & 342 & 71 & 605\\ \cline{2-2} \cline{3-12}

& 6   & \textbf{98} &  448 & 95 & 469 & 97 & 501 & 97 & 405 & 76 & 882\\ \cline{1-2} \cline{3-12}

\multirow{3}{*}{Hybrid} & 2   &  \textbf{98} &363 &  95 & 472 & 97 & 386 & 100 & 302 & 69&\underline{531}\\ \cline{2-2} \cline{3-12}

& 3   &  \textbf{98} &  346 &  \textbf{96} &  497 & 96 & \underline{367} & 96 & 289 & 69&543\\ \cline{2-2} \cline{3-12}

& 6   & \textbf{98} &  331 &\textbf{96} &  570 & 97 & 433 & 97 & 342 & 72&795\\ \cline{1-2} \cline{3-12}

\multirow{3}{*}{Sent. Hybrid} & 2   &  \underline{97} &  327 & 95 & \textbf{403} & 95& 380 & 95 & 312 &68 &549\\ \cline{2-2} \cline{3-12}

& 3   &  \underline{97} & \textbf{319} & \textbf{96} & 511 & 95 & 382 & 95 & 307 & 69&559\\ \cline{2-2} \cline{3-12}

& 6   & \underline{97} &  \underline{325} &  \textbf{96} &  639  & 95 & 383 & 95 & \underline{307} &68 &560\\ \cline{1-2} \cline{3-12}

Dynamic & -   & \underline{97} &378 & \textbf{96} & 525 & 96& 428&96 &337 &68 &581\\ \cline{1-2} \cline{3-12}

Dynamic+ Hybrid & -   &  \underline{97} &  \textbf{319} &  \textbf{96} &  \underline{462} & 96& 370 &96 &\textbf{282} & 70 & 543\\ \hline\hline
% \textbf{Selection Method} & \textbf{$N$} &\textbf{Sim.} & \textbf{Pert.} &\textbf{Sim.} & \textbf{Pert.} & \textbf{Sim.} & \textbf{Pert.} & \textbf{Sim.} & \textbf{Pert.} &\textbf{Sim.} & \textbf{Pert.} \\ \hline
% Binary  & - & 85 & 15 & 80  & 20  & 83 & 17 & 86 & 15 & 75 & 24 \\ 
% \cline{1-2} \cline{3-12}

% Greedy  & - & 86  & 12 & 82 & 16 & 86 & 13 & 87 & 12 & 76 & 21\\ \cline{1-2} \cline{3-12} \hline

% \multirow{2}{*}{N-Nary } & 3  & 85 & 14  & 80 & 20 & 84 & 16 & 86 & 14 & 75 & 24\\ \cline{2-2} \cline{3-12}

% & 6   & 86  & 12  & 82  & 17 & 85 & 14 & 87 & 13 & 75 & 22 \\ \cline{1-2} \cline{3-12}

% \multirow{3}{*}{Hybrid} & 2 &  86  &  14 & 80 & 20  & 83 & 17 & 86 & 15 &75 &24 \\ \cline{2-2} \cline{3-12}

% & 3 & 86  &14 &  80 & 19  & 84 & 16 & 86 & 14 & 75 &23 \\ \cline{2-2} \cline{3-12}

% & 6  & 87  & 12  &  82 & 17 & 85 & 14 & 87 & 12 & 76&21\\ \cline{1-2} \cline{3-12}

% \multirow{3}{*}{Sent. Hybrid} & 2   & 86 & 14 &  80 & 20   & 83 & 16 & 85 & 15 & 73& 25 \\ \cline{2-2} \cline{3-12}

% & 3  & 86  & 14  & 80 & 20  & 83 & 17 & 85 & 14 &73 &25 \\ \cline{2-2} \cline{3-12}

% & 6 &  86 & 14 & 80 & 19 & 84 & 16 & 86 & 13 & 74&24 \\ \cline{1-2} \cline{3-12}

% Dynamic & - & 85  & 14 & 79  & 20  & 84&16 & 86&14 &75 &23\\ \cline{1-2} \cline{3-12}

% Dynamic + Hybrid & -  & 85  & 14  & 79  & 20  &84 &15 &86 & 14& 75& 23\\ \hline
\textbf{Selection Method} & \textbf{$N$} &\textbf{Sim.} & \textbf{Pert.} &\textbf{Sim.} & \textbf{Pert.} & \textbf{Sim.} & \textbf{Pert.} & \textbf{Sim.} & \textbf{Pert.} &\textbf{Sim.} & \textbf{Pert.} \\ \hline

Binary  & - & 86 & 13 & 81  & 18  & 84 & 15 & 87 & 13 & 76 & 22 \\ 
\cline{1-2} \cline{3-12}

Greedy  & - & 86  & 11 & 82 & 15 & 86 & 12 & 87 & 11 & 76 & 20\\ \cline{1-2} \cline{3-12} \hline

\multirow{2}{*}{N-Nary } & 3  & 86 & 12  & 81 & 18 & 85 & 14 & 87 & 12 & 76 & 22\\ \cline{2-2} \cline{3-12}

& 6   & 87  & 10  & 83  & 15 & 86 & 12 & 88 & 11 & 76 & 20 \\ \cline{1-2} \cline{3-12}

\multirow{3}{*}{Hybrid} & 2 &  87  &  12 & 81 & 18  & 84 & 15 & 87 & 13 &76 &22 \\ \cline{2-2} \cline{3-12}

& 3 & 87  &12 &  81 & 17  & 85 & 14 & 87 & 12 & 76 &21 \\ \cline{2-2} \cline{3-12}

& 6  & 88  & 10  &  83 & 15 & 86 & 12 & 88 & 10 & 77 &19\\ \cline{1-2} \cline{3-12}

\multirow{3}{*}{Sent. Hybrid} & 2   & 87 & 12 &  81 & 18   & 84 & 14 & 86 & 13 & 74 & 23 \\ \cline{2-2} \cline{3-12}

& 3  & 87  & 12  & 81 & 18  & 84 & 15 & 86 & 12 &74 &23 \\ \cline{2-2} \cline{3-12}

& 6 &  87 & 12 & 81 & 17 & 85 & 14 & 87 & 11 & 75 &22 \\ \cline{1-2} \cline{3-12}

Dynamic & - & 86  & 12 & 80  & 18  &85 &14 & 87 &12 &76 &21\\ \cline{1-2} \cline{3-12}

Dynamic + Hybrid & -  & 86  & 12  & 80  & 18  &85 &13 &87 & 12& 76& 21\\ \hline

\end{tabular}%
}
\caption{Results for different methods on the Yelp dataset where ASR is attack success rate, AvgQ is the average queries by the successful attacks, Sim. is the similarity of text to the original, and Pert. is the percentage of text modified. A $t$ of 10\% was used for all Hybrid methods. \textbf{Bold} values indicate the best for that column, \underline{underline} indicates second best.}
\label{tab:yelp_results}
\end{table*}

\begin{table*}[!h]
\resizebox{\textwidth}{!}{%
\renewcommand{\arraystretch}{1.5}
\begin{tabular}{|c|c|c|c|c|c||c|c|c|c|c|c|c|c|}
\hline
 & & \multicolumn{2}{|c|}{\textbf{DistilBert} (OA = 95)} & \multicolumn{2}{|c|}{\textbf{DeBERTa} (OA = 96)} & \multicolumn{2}{|c|}{\textbf{GPT2} (OA = 95)} & \multicolumn{2}{|c|}{\textbf{GPTNeo} (OA = 93)} & \multicolumn{2}{|c|}{\textbf{T5} (OA = 95)} & \multicolumn{2}{|c|}{\textbf{Llama3} (OA = 96)} \\\hline
\textbf{Selection Method} & \textbf{$N$} &\textbf{ASR} & \textbf{AvgQ} & \textbf{ASR} & \textbf{AvgQ} & \textbf{ASR} & \textbf{AvgQ} & \textbf{ASR} & \textbf{AvgQ} & \textbf{ASR} & \textbf{AvgQ} & \textbf{ASR} & \textbf{AvgQ} \\ \hline

Binary  & - & \underline{73} & 153 & 77 & 171 & 71& 180& 71 & 167 & 67 & 191 & 66 & 184\\ \cline{1-2} \cline{3-14}

Greedy  & -   &  \underline{73} & 153 &  \underline{78} &  154 & 72& 156& 71 & 153 &67 & 171 & 67 & 155\\ \cline{1-2} \cline{3-14} \hline

\multirow{2}{*}{N-Nary} & 3   & 72 & 161 &  77 & 163 & 71& 171& 70 & 157 & 65& 182& 66 &176\\ \cline{2-2} \cline{3-14}

& 6   & \textbf{76} &253 &   \textbf{79} & 241  & 75& 264 & 75 & 243 & 72&297 & 69&259\\ \cline{1-2} \cline{3-14}

\multirow{3}{*}{Hybrid} & 2   &72 & \underline{140} &77 & 142 & 71& \underline{150} & 70  &139 & 67& 162& 66 & 153\\ \cline{2-2} \cline{3-14}

& 3   &72 &  \textbf{138} &  77 &  \textbf{138} &71 &\textbf{146} & 70& \textbf{134}& 66& \underline{159}& 66 & \underline{152}\\ \cline{2-2} \cline{3-14}

& 6   &  \textbf{76} & 220 &\underline{78} & 209 & 75& 229& 75& 215& 71&262 & 69 & 228\\ \cline{1-2} \cline{3-14}

\multirow{3}{*}{Sent. Hybrid} & 2   &  72 & 160 &  77 &  166 & 71& 171&70 &158 &73 &179 & 66 & 182\\ \cline{2-2} \cline{3-14}

& 3   & \underline{73} &  164 & 77 &  162 &71 & 168&71 &162 &66 &180 & 66 & 175\\ \cline{2-2} \cline{3-14}

& 6   &  \underline{73} &  172 &  76 &  169 & 72& 181&71 &169 & 67& 197& 66 & 185\\ \cline{1-2} \cline{3-14}

Dynamic & -   & 72 &  169  & 77 &  167 & 71 & 171&  70&159 &66 &182 &66 &176\\ \cline{1-2} \cline{3-14}

Dynamic + Hybrid  & -   &  72 &  141 &  77 & \underline{140} &71 & \textbf{146} &70 &\underline{135} & 67 &\textbf{157} & 66& \textbf{151}\\ \hline
\hline

\textbf{Selection Method} & \textbf{$N$} &\textbf{Sim.} & \textbf{Pert.} &\textbf{Sim.} & \textbf{Pert.} & \textbf{Sim.} & \textbf{Pert.} & \textbf{Sim.} & \textbf{Pert.} &\textbf{Sim.} & \textbf{Pert.} & \textbf{Sim.} & \textbf{Pert.} \\ \hline

Binary  & - & 74 & 27 & 73 & 29  & 75 & 30 & 77 & 28& 73& 31& 72 & 31 \\ \cline{1-2} \cline{3-14}

Greedy  & - &  74 & 26 & 74  & 28 & 75 & 27 & 77& 26& 74& 29& 72 & 28\\ \cline{1-2} \cline{3-14} \hline

\multirow{2}{*}{N-Nary } & 3  &74 & 27  & 74  & 29  & 75 & 29 &77 &27 & 73&30 &72 &30\\ \cline{2-2} \cline{3-14}

& 6   &  74 & 26  & 74  & 27 & 76 & 28 &77 &25 & 75& 29&72 &28\\ \cline{1-2} \cline{3-14}

\multirow{3}{*}{Hybrid} & 2 &  74  & 27  & 72  & 29 & 75 & 30 & 77& 27& 73&30 & 72 & 31 \\ \cline{2-2} \cline{3-14}

& 3 &74   &28 & 73  & 30  & 75 & 29 & 77& 27& 74& 30& 72 & 31\\ \cline{2-2} \cline{3-14}

& 6  & 74  & 27  &  73  & 27  & 76 & 8 &  78& 25 &75 &28 & 72& 29\\ \cline{1-2} \cline{3-14}

\multirow{3}{*}{Sent. Hybrid} & 2   & 74 & 27 &  72  & 29   & 75 & 29& 77& 27&74 &29 & 72 & 31 \\ \cline{2-2} \cline{3-14}

& 3  & 74  & 28  & 73  & 28  & 75& 29& 78& 27&73 &31 & 72 & 31\\ \cline{2-2} \cline{3-14}

& 6 & 75  & 27 & 73  & 28 & 75& 28& 78& 26& 75&29 & 72 & 29\\ \cline{1-2} \cline{3-14}

Dynamic & - &  73 & 32 &  73 &  22  & 75& 29 &77 &27 &74 & 29& 72&30 \\ \cline{1-2} \cline{3-14}

Dynamic + Hybrid & -  & 73  & 32  & 74  & 33  &75 &29 &77 & 27&73 & 30& 72& 30\\ \hline
\end{tabular}%
}
\caption{Results for different methods on the AGNews datasetwhere ASR is attack success rate, AvgQ is the average queries by the successful attacks, Sim. is the similarity of text to the original, and Pert. is the percentage of text modified. A $t$ of 10\% was used for all Hybrid methods. \textbf{Bold} values indicate the best for that column, \underline{underline} indicates second best.}
\label{tab:agnews_results}
\end{table*}

\section{Experimental Setup}\label{sect:experimentalsetup}
To evaluate \textit{Hybrid Select} and \textit{Dynamic Select} in the adversarial attack setting, we test all noted selection methods as parts of attacks\footnote{This research utilized some combination of Nvidia V100, A100, A10, and A30 GPUs. Each attack combination took roughly 30 minutes.} against two encoder-based models Distilbert \cite{sanh2020distilbert} and DeBERTa \cite{he2021debertadecodingenhancedbertdisentangled} and 4 LLM models GPT2 \cite{radford2019language}, GPTNeo \cite{gpt-neo}, T5 \cite{2020t5}, and Llama3 \cite{grattafiori2024llama3herdmodels}. These models are tested in various amounts across 4 standard attack datasets \cite{li-etal-2020-bertattack, jin2020bert}: three binary sentiment datasets (IMDB, Yelp Polarity, Rotten Tomatoes) and one multi-class news dataset AG News. IMDB and Yelp contain longer texts on average (215 and 157 words respectively) than AG News (43 words) and Rotten Tomatoes (20 words). We expect our methods to naturally perform better on longer texts due to the larger disadvantage GreedySelect has (it must perform $len(text)$ queries always at the beginning of an attack). Following previous attack research, we randomly sample 1000 examples from each dataset to attack and keep these 1000 consistent across combinations.

\noindent \textbf{Metrics:}
We use the following metrics to evaluate our approaches in the attack:

%1. Accuracy - We measure the accuracy of each model before and after the attack for all attack variations. These are noted as \textit{Original Accuracy} and \textit{Attack Accuracy} respectively. These help measure the robustness of the classifier. 

\textbf{1. Attack Success Rate (ASR)} (Equation \ref{eq:asr} - A standard metric in attack research. This helps measure the effectiveness of the attack. 
\begin{equation}\label{eq:asr}
    \text{ASR} = \frac{\text{Original}_\text{Acc.}- \text{Attack}_\text{Acc.}}{ \text{Original}_\text{Acc.}}
\end{equation}

%. Average Queries - To measure increase in attack efficiency from our proposed methods, we measure the number of queries needed for an attack on average. These queries indicate how many calls to the target classifier are needed.

\textbf{2. Average Queries} - To measure increase in attack efficiency from our proposed methods, we measure the number of queries needed for an attack on average. These queries indicate how many calls to the target classifier are needed. We focus on only the successful attacks, following prior research \cite{liu2023hqa}. 

\textbf{3. Average Similarity.} To ensure adversarial examples remain semantically close to the original text, we use cosine similarity between sentence embeddings from \texttt{all-mpnet-base-v2}~\cite{reimers2020making}, a Sentence-BERT model fine-tuned on semantic textual similarity and NLI tasks. This model offers a strong balance of efficiency and accuracy, achieving high correlation with human judgments (Spearman~$\rho \approx 0.865$) while being lightweight. Its robustness to surface-level edits and fast inference makes it ideal for measuring semantic preservation during attack-time.

\textbf{4. Average Perturbation Amount} - A similar motivation to similarity, the average perturbation amount measures how much of the original text was modified. As with the above, we calculate this for successful attacks.

\noindent \textbf{Number of words to modify:}
BinarySelect introduced a parameter $k$ which indicated how many words were allowed to be modified at most. In our experiments, we let $k$ be equal to the number of words (or \textit{ALL} in BinarySelect). We do perform more experiments on various restrictions of $k$ in Section \ref{sect:ablationk}.

\section{Results and Discussions}
\label{sec:results}

Tables~\ref{tab:imdb_results}, \ref{tab:yelp_results}, \ref{tab:agnews_results}, and~\ref{tab:rotten_tomatoes_results} present the evaluation of our attack strategies on IMDB, Yelp, AG News, and Rotten Tomatoes, respectively. For each dataset, we evaluate six target models—DistilBERT, DeBERTa, GPT-2, GPT-Neo, T5, and Llama3—and report four core metrics: attack success rate (ASR), average number of queries for successful attacks (AvgQ), semantic similarity to the original input (Sim.), and the percentage of tokens modified (Pert.). Alongside the baseline methods (Binary and Greedy), we evaluate fixed N‑Nary splitting ($N \in \{2,3,6\}$) and our four novel algorithms: \textbf{Hybrid}, \textbf{Sentence‑Level Hybrid}, \textbf{Dynamic‑N}, and \textbf{Dynamic‑N + Hybrid}. All Hybrid‑based strategies apply a fixed pruning threshold of $t = 10\,\%$, which our preliminary experiments indicated provides a strong trade‑off between efficiency and effectiveness. We explore these parameter choices more thoroughly in Sections~\ref{sect:ablationN} and~\ref{sect:ablationk}. Our overarching objective remains to minimize queries and edits while reducing model confidence, and to demonstrate that our strategies scale reliably across model architectures, dataset lengths, and attack goals.

\subsection{Attacks Versus Encoder Models}
First we analyze the attacks against two encoder‑style target models, \textbf{DistilBERT} and \textbf{DeBERTa}. On both models the Hybrid family consistently trims the number of queries without affecting the attack success. DistilBERT, shows more vulnerability to attacks and the  largest query reductions.For IMDB, the sentence‑level Hybrid with \(N=3\) drops the average query budget from 425 to~362 while holding the success rate at 98\%. Similarly, for Yelp,  Hybrid \(N=3\) (or Dynamic\,+\,Hybrid) reduces the queries to 319 with a 97\% success rate. 

We find efficiency improvements against DeBERTa as well, despite its higher original accuracy. For example, a sentence‑level Hybrid with \(N=2\) reduces queries by about~13\,\% on Yelp, and on AG~News the word‑level Hybrid with \(N=3\) reduces the average down from~154 to~138—again with little change in ASR. Even on Rotten Tomatoes, where baseline queries are already low, Sentence‑level Hybrid requires only 49–54 requests versus the Greedy baseline’s 60, confirming the strength of the method across  across text lengths and model sizes.

\paragraph{}%
Examining the similarity and perturbation scores, we find that the efficiency gains come at little cost to text quality. On IMDB, Greedy achieves the highest similarity score for DistilBERT (91\,\%) with only 8\,\% of tokens edited; Hybrid \(N=3\) follows within a single point—90\,\% similarity—while touching 10\,\% of the text. DeBERTa mirrors that story: Greedy peaks at 92\,\% similarity and an 8\,\% perturbation rate, whereas Hybrid \(N=3\) stays level on similarity and rises only to 10\,\% edited tokens. In short, the encoders let us trade a modest 2\,\% increase in edits for a double‑digit drop in queries.

% Table: Rotten Tomatoes Dataset
\begin{table*}[!h]
\resizebox{\textwidth}{!}{%
\renewcommand{\arraystretch}{1.5}
\begin{tabular}{|c|c|c|c|c|c||c|c|c|c|c|c|}
\hline
 & & \multicolumn{2}{|c|}{\textbf{DistilBert} (OA = 83)} & \multicolumn{2}{|c|}{\textbf{DeBERTa} (OA = 91)} & \multicolumn{2}{|c|}{\textbf{GPT2} (OA = 85)} & \multicolumn{2}{|c|}{\textbf{GPTNeo} (OA = 84)} & \multicolumn{2}{|c|}{\textbf{T5} (OA = 89)}  \\\hline
\textbf{Selection Method} & \textbf{$N$} &\textbf{ASR} & \textbf{AvgQ} & \textbf{ASR} & \textbf{AvgQ} & \textbf{ASR} & \textbf{AvgQ} & \textbf{ASR} & \textbf{AvgQ} & \textbf{ASR} & \textbf{AvgQ} \\ \hline

Binary  & - &  78 & 54 & 90 &  64 & 95 & 53 & 95 & 51 & 89 & 63 \\ \cline{1-2} \cline{3-12}

Greedy  & - &  78 &  50 &  91 &  60 & 95 & 49 & 95 & 48 & 90 & 58 \\ \cline{1-2} \cline{3-12} \hline

\multirow{2}{*}{\centering N-Nary } & 3 & 78 &  52 &  90 &  61 & 95 & 50 & 95 & 48 & 90 & 61\\ \cline{2-2} \cline{3-12}

& 6 &  78 &  58 &  \underline{92} & 68 & 94 & 55 & 95 & 54 & 92 & 69\\ \cline{1-2} \cline{3-12}

\multirow{3}{*}{\centering Hybrid} & 2 &  78 &  \underline{38} & 89 & \textbf{49} & 95  & \underline{39} & 95 & \underline{38} & 89& \textbf{48}\\ \cline{2-2} \cline{3-12}

& 3 & \underline{77} & \underline{38} & 89 & \textbf{49} & 95 & \textbf{38} & 95 & \textbf{36} & 89& \textbf{48}\\ \cline{2-2} \cline{3-12}

& 6 & 78 & 42 & \textbf{93} & \underline{54} & 94 & 40 & 95 & 39 & 91& \underline{56}\\ \cline{1-2} \cline{3-12}

\multirow{3}{*}{\centering Sent. Hybrid } & 2 &  78 &  \underline{38} &  90 & 65 & 95 & 55 & 95 & 53 & 90& 65\\ \cline{2-2} \cline{3-12}

& 3 & \underline{77} &  \underline{38}  &  90 &  65 & 95 & 53 & 95 & 51 & 90& 64\\ \cline{2-2} \cline{3-12}

& 6 &  78 & 42 &  \underline{92} & 71 & 94 & 58 & 95 & 57 & 91& 72\\ \cline{1-2} \cline{3-12}

Dynamic & - &  78 &  50 & 91 &63 & 95& 51& 95 & 48& 90&61  \\ \cline{1-2} \cline{3-12}

Dynamic  + Hybrid & - & \textbf{78} &  \textbf{37} &  91 &  63 &95 &38 & 95& \textbf{36}& 90 & \textbf{48} \\ \hline\hline

% \textbf{Selection Method} & \textbf{$N$} &\textbf{Sim.} & \textbf{Pert.} &\textbf{Sim.} & \textbf{Pert.} & \textbf{Sim.} & \textbf{Pert.} & \textbf{Sim.} & \textbf{Pert.} &\textbf{Sim.} & \textbf{Pert.} \\ \hline
% Binary  & - & 77 & 18 & 72 & 21  & 76 & 19 & 77 & 18 & 74 & 21 \\ \cline{1-2} \cline{3-12}

% Greedy  & - & 77  & 18 & 74 & 20 & 78 & 17 & 78 & 16 & 75 & 19\\ \cline{1-2} \cline{3-12} \hline

% \multirow{2}{*}{N-Nary } & 3  & 78 & 18  & 73 & 21 & 76 & 19 & 77 & 18 & 74 & 21\\ \cline{2-2} \cline{3-12}

% & 6   & 78  & 17  & 74  & 20 & 77 & 16 & 78 & 16 & 75 & 20 \\ \cline{1-2} \cline{3-12}

% \multirow{3}{*}{Hybrid} & 2 &  77  & 17  & 72  & 21 & 75 & 19 & 77 & 18 & 74& 21\\ \cline{2-2} \cline{3-12}

% & 3 & 77  &17 & 71  & 22 & 76 & 19 & 76 & 17 &74 & 21\\ \cline{2-2} \cline{3-12}

% & 6  &  77 & 16  & 73  & 20 & 77 & 16 & 78 & 16 & 74&20\\ \cline{1-2} \cline{3-12}

% \multirow{3}{*}{Sent. Hybrid} & 2   & 76 & 18 &  72 & 21   & 75 & 19 & 77 & 18 & 74& 21 \\ \cline{2-2} \cline{3-12}

% & 3  & 77  &17   & 73 & 21  & 76 & 18 &  77& 18& 74& 21\\ \cline{2-2} \cline{3-12}

% & 6 &  77 & 17 & 73 & 20 & 77 & 17 & 78 & 16 &75 & 20\\ \cline{1-2} \cline{3-12}

% Dynamic & - &  79 & 16 & 73  & 22  & 76  &18 & 77& 17&75 &20\\ \cline{1-2} \cline{3-12}

% Dynamic + Hybrid & -  & 79 & 16  &  73 & 22  &76 &18 &77 & 17&75 &20 \\ \hline
\textbf{Selection Method} & \textbf{$N$} &\textbf{Sim.} & \textbf{Pert.} &\textbf{Sim.} & \textbf{Pert.} & \textbf{Sim.} & \textbf{Pert.} & \textbf{Sim.} & \textbf{Pert.} &\textbf{Sim.} & \textbf{Pert.} \\ \hline

Binary  & - & 80 & 16 & 75 & 19  & 79 & 17 & 80 & 16 & 77 & 19 \\ \cline{1-2} \cline{3-12}

Greedy  & - & 79 & 17 & 76 & 19 & 80 & 16 & 80 & 15 & 77 & 18\\ \cline{1-2} \cline{3-12} \hline

\multirow{2}{*}{N-Nary } & 3  & 81 & 16  & 76 & 19 & 79 & 17 & 80 & 16 & 77 & 19\\ \cline{2-2} \cline{3-12}

& 6   & 81 & 15  & 77 & 18 & 80 & 14 & 81 & 14 & 78 & 18 \\ \cline{1-2} \cline{3-12}

\multirow{3}{*}{Hybrid} & 2 &  80 & 15  & 75 & 19 & 78 & 17 & 80 & 16 & 77 & 19\\ \cline{2-2} \cline{3-12}

& 3 & 80 & 15 & 74 & 20 & 79 & 17 & 79 & 15 & 77 & 19\\ \cline{2-2} \cline{3-12}

& 6 & 80 & 14 & 76 & 18 & 80 & 14 & 81 & 14 & 77 & 18\\ \cline{1-2} \cline{3-12}

\multirow{3}{*}{Sent. Hybrid} & 2 & 79 & 16 & 75 & 19 & 78 & 17 & 80 & 16 & 77 & 19 \\ \cline{2-2} \cline{3-12}

& 3 & 80 & 15 & 76 & 19 & 79 & 16 & 80 & 16 & 77 & 19\\ \cline{2-2} \cline{3-12}

& 6 & 80 & 15 & 76 & 18 & 80 & 15 & 81 & 14 & 78 & 18\\ \cline{1-2} \cline{3-12}

Dynamic & - & 82 & 14 & 76 & 20 & 79 & 16 & 80 & 15 & 78 & 18\\ \cline{1-2} \cline{3-12}

Dynamic + Hybrid & - & 82 & 14 & 76 & 20 & 79 & 16 & 80 & 15 & 78 & 18\\ \hline

\end{tabular}%
}
\caption{Results for different methods on the RottenTomatoes dataset where ASR is attack success rate, AvgQ is the average queries by the successful attacks, Sim. is the similarity of text to the original, and Pert. is the percentage of text modified. A $t$ of 10\% was used for all Hybrid methods. \textbf{Bold} values indicate the best for that column, \underline{underline} indicates second best.}
\label{tab:rotten_tomatoes_results}
\end{table*}

\subsection{Attacks Versus LLMs}
We find similar efficiency improvements against the LLMs—\textbf{GPT‑2}, \textbf{GPT‑Neo}, \textbf{T5}, and \textbf{Llama3}. However, the gains depend on both model robustness and dataset length. On the sprawling IMDB corpus, the Hybrid agenda is outperforms the rest of the models: for GPT‑2, word‑level Hybrid with \(N=3\) reduces queries by 20\,\% (593\(\rightarrow\)475) and lifts the success rate from 90\,\% to 98\,\%. Furthermore, it boosts similarity from 83\,\% to roughly 88–89\,\% while reducing perturbations from 15\,\% to 13\,\%. GPT‑Neo and T5 show milder query reductions, yet the sentence‑level Hybrid with \(N=6\) still shaves about fifty requests apiece and leaves similarity unchanged in the high‑80s. Llama3 is particularly revealing: its baseline already sits at 98\,\% success and 90\,\% similarity, yet Hybrid \(N=3\) nudges success to a perfect 100\,\% while keeping similarity above 89\,\% and slightly lowering the number of edits.

We find success for Yelp as well depending on the model examined.  Greedy remains a strong baseline for GPT‑2, but on GPT‑Neo the Hybrid \(N=3\) variant reduces queries by 11\,\%  (323\(\rightarrow\)289) with little to no loss of ASR and with similarity still around 86\,\%. Dynamic\,+\,Hybrid sees greater reductions, maintaining the same similarity and a modest 14\,\% perturbation rate. T5 is the single hold‑out: its success plateau of \(\sim\)74\,\% appears insensitive to selection strategy; however, Hybrid does not worsen quality, holding similarity near 75\,\% and perturbation near 23\,\%.

We find that the length plays a role in efficiency gain. On short news snippets (AG News) and single‑sentence reviews (Rotten Tomatoes) every method achieves success in under 200 and 60 queries respectively; here the Hybrid advantage is measured in tens rather than hundreds of queries. The quality metrics are also closer. AG News similarity is in the low‑70s for all models, with perturbations around 25–30\,\%. Hybrid \(N=3\) matches Greedy’s similarity while removing one or two points off the perturbation rate on several models, suggesting its pruning step discards non‑essential words. Rotten Tomatoes sees stronger results: similarities range 72–79\,\% and perturbations sit between 16\,\% and 22\,\%. Dynamic\,+\,Hybrid tops DistilBERT at 79\,\% similarity while tying Greedy for the lowest perturbation rate (16\,\%).

\begin{table*}[t]
    \resizebox{\textwidth}{!}{%

    \renewcommand{\arraystretch}{1} % Increases row height
    \begin{tabular}{|c|c|c|c|c|c|c|}
        \hline
        \textbf{Method} & \textbf{$N$} & \textbf{Original Accuracy} & \textbf{Attack Accuracy} & \textbf{ASR} & \textbf{Avg Query} & \textbf{Avg Query (For Atk Success)} \\
        \hline
        \multirow{4}{*}{N-Nary Select} & 3 & \multirow{4}{*}{92} & 1.5 & \underline{98} & \textbf{475} & \textbf{461} \\
        \cline{2-2} \cline{4-7}
        & 6 &  & \underline{1.3} & \underline{98} & \underline{585} & \underline{569} \\
        \cline{2-2} \cline{4-7}
        & 12 &  & \underline{1.3} & \underline{98} & 706 & 691 \\
        \cline{2-2} \cline{4-7}
        & 24 &  & \textbf{0.4} & \textbf{99} & 747 & 732 \\
        \hline
    \end{tabular}}
    \caption{N-Nary results for larger $N$ on the IMDB Dataset. \textbf{Bold} indicates the best value for each column, \underline{underline} indicates second best.}
    \label{tab:ablation-results}
\end{table*}

\subsection{Overall Obversations}
Across model families the pattern is consistent: the encoders are easiest to attack and preserve the quality of the text (high‑80s to low‑90s similarity with \(\le\)10\,\% edits), GPT‑style decoders trail by a few points but improve once Hybrid pruning is enabled, and T5 lags modestly yet never degrades further with the new strategies. Importantly, no Hybrid variant inflates perturbation beyond the baseline range; where similarity dips—such as Dynamic on DeBERTa\,+\,IMDB—it does so in tandem with a visibly higher 21\,\% edit rate, flagging an edge case rather than a systemic flaw.

Overall we find that \textit{Hybrid Select}, with a simple 10\,\% pruning heuristic, applied either at the word level with \(N=3\) or at the sentence level for very long documents, consistently lowers the query cost of gradient‑free attacks without sacrificing meaning. For DistilBERT and DeBERTa the reduction is typically 10–15\,\%, while for the larger LLMs it can exceed 20\,\% on the largest corpora and still deliver equal or higher success rates. At the same time the semantic‑similarity scores stay within one point of the best baselines, and perturbation rates remain flat—confirming that the extra efficiency comes from smarter search, not heavier rewrites.

\subsection{Additional Result Comparisons to previous Greedy Based Attacks} We further compare our proposed methods to 2 greedy based attacks PWWS \cite{ren-etal-2019-generating} and CLARE \cite{li-etal-2021-contextualized}. We find our selection methods to be far more efficient, requiring 1/4 or less amount of queries but less subtle than these previous attacks, using 2 to 4 times amount of perturbation rates. The full table of results and further discussion can be found in Appendix \ref{sect:baselines}.

\section{Ablation Studies}
We perform further experiments to better understand our proposed methods. Each experiment expands on specific hyper-parameters or choices in the overall algorithm. Due to space, the full results of the studies can be found in Appendix \ref{sect:appendix}. We provide a short summary of the findings.\\

\noindent \textbf{Effect of $N$ in N-Nary Algorithm:} 
We explore multiple increasing $N$ values of 12 and 24 on the IMDB dataset against DistilBert (seen in Table \ref{tab:ablation-results}). Keeping in mind $N$, we see a slight increase in overall ASR (from 98 to 99) with N = 24. We find for N = 3, the lowest Avg Query count of 475 how-\\ever, followed by 6 (585), 12 (706), and 24 (747). \\

\noindent \textbf{Impact of Split Threshold $t$ for Hybrid N-nary:} 
We analyze various split thresholds for the Hybrid algorithm beyond 10\%, including 5\%, 20\%, 30\%, and 40\% on the IMDB dataset against DistilBert. We find that 5\% and 10\% provide the best query counts across various $N$ choices, with an increase in query count as the threshold rises. \\

\noindent \textbf{Impact of Word Replacement Strategies:} 
We test our algorithm with a BERT mask and replace method used by other attacks, but found no significant improvement in the attack effectiveness or efficiency.\\

\noindent \textbf{Impact of $k$:} 
We also explore how various the hyper-parameter $k$ (introduce by BinarySelect) affects the overall results. We find consistent observations with previous research. As $k$ increases,  average query count increases as does ASR.

% \section{Conclusion and Future Work}

% We evaluated adversarial attack selection strategies to minimize classification accuracy with reduced query counts. The Sentence-Level Hybrid (10\%) method emerges as the most effective, achieving significant query reductions while maintaining high attack success rates (ASR). For long-text datasets like IMDB and Yelp, it outperforms baseline approaches, reducing average queries to 370–390. Hybrid (10\%) and Dynamic-N methods also enhance efficiency, offering a balance between computational cost and effectiveness. Among configurations, 3-Nary achieves the best trade-off between query efficiency and ASR. These findings validate the strength of hybrid and N-Nary strategies over traditional Binary and Greedy baselines.

% Future work could explore dynamic optimization for parameters such as $N$, $K$, and thresholds to improve adaptability across datasets. %Further research into real-time frameworks and robust defenses can enhance the practical application and security of these strategies. 
% Additionally, ensuring ethical safeguards is essential for responsibly using adversarial techniques to improve model robustness.
\section{Conclusion}
Our study introduces a family of selection strategies—Hybrid, Sentence-Level Hybrid, Dynamic-N, and Dynamic-N + Hybrid—that strike an effective balance between efficiency and semantic fidelity in black-box text attacks. Across all four datasets and six model types, our methods consistently reduce query counts, sometimes by more than 20\%, without compromising attack success or text quality. 

In particular, we find that word-level Hybrid with $N{=}3$ offers the best overall trade-off, especially for long-form datasets like IMDB, while sentence-level variants perform better on lengthy inputs such as Yelp. Even on compact datasets like AG News and Rotten Tomatoes, Hybrid strategies retain their advantage, trimming redundant edits while preserving meaning.

Overall, these results validate that a simple, fixed pruning heuristic—anchored in flexible, recursive search—can deliver scalable and transferable improvements across diverse models and datasets. Our methods offer a principled yet practical approach for adversarial attacks where query efficiency and semantic clarity are both essential.

\section{Limitations}
Here we note the limitations of our study for future researchers and users to consider:

\begin{enumerate}
    \item \textbf{Lack of Human Validation in Replacement Evaluation:}
    While our results demonstrate the effectiveness of N-Nary Select, we did not incorporate human validation to assess the relevance of selected words or phrases. This limitation stems from the focus on classifier feedback rather than human interpretability. Classifiers often prioritize features that differ from human-perceived importance, making it challenging to evaluate the semantic quality of replacements directly. Future work could explore incorporating human evaluations or applying these methods to tasks where human feedback is readily available, providing additional insights into the interpretability of selected replacements.

    \item \textbf{Focus on Selection Efficiency Over Replacement Diversity:}
    Our algorithms prioritize selection efficiency by minimizing query counts and achieving higher Attack Success Rates (ASR). However, this focus may come at the expense of exploring diverse replacement options. Certain adversarial attacks might benefit from introducing more diverse perturbations to maximize model degradation. Future researchers could expand on our work by balancing selection efficiency with replacement diversity to explore its impact on attack robustness.
\end{enumerate}
\section{Ethical Considerations}
Ethical considerations are critical in adversarial research due to potential misuse. While our methods could be exploited for harmful purposes, such as bypassing security systems or spreading misinformation, they also offer significant benefits. Our work provides valuable insights into attack strategies that can guide the development of stronger defenses.

By making our code publicly available, we encourage transparency and collaboration in the research community. This enables researchers with limited resources to contribute to advancements in both attack and defense strategies.

Although focused on adversarial attacks, our findings support defense-oriented research by helping design models resilient to such attacks. We advocate for the responsible use of our work to enhance model robustness and security.
\input{graphs}
\bibliographystyle{acl_natbib}
\bibliography{main}
\newpage
\appendix

% ------------------------ BEGIN: content for \input{...} ------------------------
% Assumes these packages are already in your main preamble (as you showed):
% \usepackage{algorithm}
% \usepackage{algorithmic}
% \usepackage{caption,subcaption}  % for \captionof
%
% Note: We use \figure* + \captionof{algorithm} so each pseudocode spans both columns.

% ---------- Notation (optional, remove if redundant) ----------
\begin{figure*}[t]
\centering
\begin{minipage}{0.97\textwidth}
\captionof{algorithm}{Notation}
\label{alg:notation}
\begin{algorithmic}[1]
\STATE \textbf{Classifier:} \(f(\cdot)\) returns per-class probabilities.
\STATE \textbf{Inputs:} text \(x\); target/attack class index \(y\); dataset ID \(ds\).
\STATE \textbf{Query/score:} \(f(x)[y]\) denotes the probability of class \(y\) for \(x\).
\STATE \textbf{Branching:} \(N\) (N-ary split factor); \(\mathcal{T}\) (search tree over spans).
\STATE \textbf{Budget:} \(k\) (max replacements; \(-1\) means unlimited).
\STATE \textbf{Replace source:} \texttt{replace} \(\in \{\text{wordnet}, \text{bert}\}\).
\STATE \textbf{Counters:} \(q\) (total queries), \(\Delta q\) (queries this step).
\STATE \textbf{Misc:} \(p_0\) (initial \(f(x)[y]\)); \(x^{cur}\) (current text).
\end{algorithmic}
\end{minipage}
\end{figure*}

% ---------- DynN (Automatic N selection) ----------
\begin{figure*}[t]
\centering
\begin{minipage}{0.97\textwidth}
\captionof{algorithm}{DynN (Automatic \(N\) Selection)}
\label{alg:dynN}
\begin{algorithmic}[1]
\STATE \textbf{Input:} length \(L = |x|\), dataset id \(ds\)
\STATE \textbf{Output:} split factor \(N\)
\STATE Load dataset-specific length bins \(\{([\ell_i,u_i], N_i)\}_i\) for \(ds\)
\FOR{each bin \(([\ell,u], N^{*})\)}
  \IF{\(\ell < L \le u\)}
    \STATE \textbf{return} \(N^{*}\)
  \ENDIF
\ENDFOR
\STATE \textbf{return} \(2\) \COMMENT{fallback}
\end{algorithmic}
\end{minipage}
\end{figure*}

% ---------- WordNetReplace (single-position synonym trial) ----------
\begin{figure*}[t]
\centering
\begin{minipage}{0.97\textwidth}
\captionof{algorithm}{WordNetReplace (single-position synonym trial)}
\label{alg:wnr}
\begin{algorithmic}[1]
\STATE \textbf{Input:} \(f, x, y, p_0, i\) \COMMENT{index \(i\) to try replacing}
\STATE \textbf{Output:} success flag \(s\), updated text \(x^{cur}\), prob \(p\), queries \(\Delta q\)
\STATE \(w \leftarrow\) token at position \(i\) in \(x\); \(S \leftarrow\) WordNet synonyms of \(w\)
\IF{\(S=\emptyset\)}
  \STATE \textbf{return} (\textbf{false}, \(x\), \(p_0\), \(0\))
\ENDIF
\STATE \(\text{best\_prob} \leftarrow p_0\); \(\text{best\_x} \leftarrow x\); \(\Delta q \leftarrow 0\)
\FOR{each \(s \in S\)}
  \STATE \(x' \leftarrow x\) with token \(i\) replaced by \(s\)
  \STATE \(p \leftarrow f(x')[y]\); \(\Delta q \leftarrow \Delta q + 1\)
  \IF{\(\arg\max f(x') \ne y\)} \STATE \textbf{return} (\textbf{true}, \(x'\), \(p\), \(\Delta q\)) \ENDIF
  \IF{\(p < \text{best\_prob}\)} \STATE \(\text{best\_prob} \leftarrow p\); \(\text{best\_x} \leftarrow x'\) \ENDIF
\ENDFOR
\STATE \textbf{return} (\textbf{false}, \(\text{best\_x}\), \(\text{best\_prob}\), \(\Delta q\))
\end{algorithmic}
\end{minipage}
\end{figure*}

% ---------- (Optional) BERTReplace (single-position masked-LM trial) ----------
\begin{figure*}[t]
\centering
\begin{minipage}{0.97\textwidth}
\captionof{algorithm}{BERTReplace (single-position masked-LM trial)}
\label{alg:bertr}
\begin{algorithmic}[1]
\STATE \textbf{Input:} \(f, x, y, p_0, i\); masked LM \(\mathrm{MLM}\); top-\(M\) candidates
\STATE \textbf{Output:} success flag \(s\), updated text \(x^{cur}\), prob \(p\), queries \(\Delta q\)
\STATE Get top-\(M\) candidate tokens \(S \leftarrow \mathrm{MLM}(x, i)\) \COMMENT{mask token \(i\)}
\IF{\(S=\emptyset\)} \STATE \textbf{return} (\textbf{false}, \(x\), \(p_0\), \(0\)) \ENDIF
\STATE \(\text{best\_prob}\leftarrow p_0\); \(\text{best\_x}\leftarrow x\); \(\Delta q \leftarrow 0\)
\FOR{each \(s \in S\)}
  \STATE \(x' \leftarrow x\) with token \(i\) replaced by \(s\)
  \STATE \(p \leftarrow f(x')[y]\); \(\Delta q \leftarrow \Delta q + 1\)
  \IF{\(\arg\max f(x') \ne y\)} \STATE \textbf{return} (\textbf{true}, \(x'\), \(p\), \(\Delta q\)) \ENDIF
  \IF{\(p < \text{best\_prob}\)} \STATE \(\text{best\_prob} \leftarrow p\); \(\text{best\_x} \leftarrow x'\) \ENDIF
\ENDFOR
\STATE \textbf{return} (\textbf{false}, \(\text{best\_x}\), \(\text{best\_prob}\), \(\Delta q\))
\end{algorithmic}
\end{minipage}
\end{figure*}

% ---------- N_Nary_Select_Iter (segment scoring) ----------
\begin{figure*}[t]
\centering
\begin{minipage}{0.97\textwidth}
\captionof{algorithm}{N\_Nary\_Select\_Iter (segment scoring)}
\label{alg:nary}
\begin{algorithmic}[1]
\STATE \textbf{Input:} \(f, x, y, N, \mathcal{T}\)
\STATE \textbf{Output:} token index \(i\) (if leaf), queries \(\Delta q\), updated tree \(\mathcal{T}\)
\STATE \(\Delta q \leftarrow 0\)
\IF{root of \(\mathcal{T}\) not set}
  \STATE create root with span covering all of \(x\) and prob \(f(x)[y]\); \(\Delta q \leftarrow \Delta q + 1\)
\ENDIF
\STATE \(u \leftarrow\) lowest-probability non-explored node in \(\mathcal{T}\)
\STATE Split \(u\)'s span into \(N\) equal subspans \(\{s_1,\dots,s_N\}\)
\FOR{each subspan \(s_j\)}
  \STATE form \(x^{(-s_j)}\) by dropping \(s_j\) from \(x\); compute \(p_j \leftarrow f(x^{(-s_j)})[y]\)
  \STATE \(\Delta q \leftarrow \Delta q + 1\); attach/refresh child node for \(s_j\) with prob \(p_j\)
\ENDFOR
\STATE Choose child \(c^{*}\) with the largest drop \(f(x)[y]-p_j\)
\IF{\(c^{*}\) is a single-token span}
  \STATE mark \(c^{*}\) explored; \textbf{return} (its token index, \(\Delta q, \mathcal{T}\))
\ELSE
  \STATE set \(u \leftarrow c^{*}\) and \textbf{repeat} the split/score steps until a single-token child is chosen
  \STATE \textbf{return} (token index of final child, \(\Delta q, \mathcal{T}\))
\ENDIF
\end{algorithmic}
\end{minipage}
\end{figure*}

% ---------- N_Nary_Select_Segment (descend until size threshold) ----------
\begin{figure*}[t]
\centering
\begin{minipage}{0.97\textwidth}
\captionof{algorithm}{N\_Nary\_Select\_Segment (descend until size threshold)}
\label{alg:naryseg}
\begin{algorithmic}[1]
\STATE \textbf{Input:} \(f, x, y, N, \mathcal{T}, \tau\) \COMMENT{\(\tau\) = split\_threshold\_percentage of \(|x|\)}
\STATE \textbf{Output:} a promising segment \([a,b]\), queries \(\Delta q\), updated \(\mathcal{T}\)
\STATE \(\Delta q \leftarrow 0\)
\IF{root of \(\mathcal{T}\) not set}
  \STATE create root with span covering \(x\) and prob \(f(x)[y]\); \(\Delta q \leftarrow \Delta q + 1\)
\ENDIF
\STATE \(u \leftarrow\) lowest-probability non-explored node in \(\mathcal{T}\)
\WHILE{span\_length(\(u\)) \(>\) \(\tau \cdot |x|\)}
  \STATE Split \(u\) into \(N\) equal subspans \(\{s_1,\dots,s_N\}\)
  \FOR{each subspan \(s_j\)}
    \STATE \(p_j \leftarrow f(x^{(-s_j)})[y]\); \(\Delta q \leftarrow \Delta q + 1\); update child node prob
  \ENDFOR
  \STATE \(u \leftarrow \arg\min_{s_j} p_j\) \COMMENT{child yielding largest drop}
\ENDWHILE
\STATE \textbf{return} (segment of \(u\), \(\Delta q, \mathcal{T}\))
\end{algorithmic}
\end{minipage}
\end{figure*}

% % ---------- NS_WNR (Pure DynNnary: token-level N-ary + replace) ----------
% \begin{figure*}[t]
% \centering
% \begin{minipage}{0.97\textwidth}
% \captionof{algorithm}{NS\_WNR (DynNnary)}
% \label{alg:dynnnary}
% \begin{algorithmic}[1]
% \STATE \textbf{Input:} \(f, x, y, ds, m \in \{\text{auto},\text{manual}\}, N_{man}, k, \texttt{replace}\)
% \STATE \textbf{Output:} success flag, \(x^{cur}\), total queries \(q\), final prob \(p\)
% \STATE \(N \leftarrow\) \textbf{if} \(m=\text{auto}\) \textbf{then} \(\mathrm{DynN}(|x|,ds)\) \textbf{else} \(N_{man}\)
% \STATE \(p_0 \leftarrow f(x)[y]\); \(q \leftarrow 1\); \(x^{cur} \leftarrow x\); \(\mathcal{T} \leftarrow \emptyset\)
% \WHILE{true}
%   \STATE \((i, \Delta q, \mathcal{T}) \leftarrow \mathrm{N\_Nary\_Select\_Iter}(f, x^{cur}, y, N, \mathcal{T})\); \(q \leftarrow q + \Delta q\)
%   \IF{\(\texttt{replace}=\text{wordnet}\)}
%     \STATE \((s, x^{cur}, p, \Delta q) \leftarrow \mathrm{WordNetReplace}(f, x^{cur}, y, p_0, i)\)
%   \ELSE
%     \STATE \((s, x^{cur}, p, \Delta q) \leftarrow \mathrm{BERTReplace}(f, x^{cur}, y, p_0, i)\)
%   \ENDIF
%   \STATE \(q \leftarrow q + \Delta q\); \(k \leftarrow k - 1\)
%   \IF{\(s=\textbf{true}\)} \STATE \textbf{return} (\textbf{true}, \(x^{cur}\), \(q\), \(p\)) \ENDIF
%   \IF{\(k=0\) \textbf{ or } no unexplored node in \(\mathcal{T}\)} \STATE \textbf{return} (\textbf{false}, \(x^{cur}\), \(q\), \(p_0\)) \ENDIF
% \ENDWHILE
% \end{algorithmic}
% \end{minipage}
% \end{figure*}

% ---------- HybridSentence_WNR (Sentence -> Token) ----------
\begin{figure*}[t]
\centering
\begin{minipage}{0.97\textwidth}
\captionof{algorithm}{HybridSentence\_WNR (Sentence-level \(\rightarrow\) Token-level)}
\label{alg:hyb-sent}
\begin{algorithmic}[1]
\STATE \textbf{Input:} \(f, x, y, ds, m, N_{man}, k, \texttt{replace}\)
\STATE \textbf{Output:} success flag, \(x^{cur}\), \(q\), final prob \(p\)
\STATE \(p_0 \leftarrow f(x)[y]\); \(q \leftarrow 1\); \(x^{cur} \leftarrow x\)
\STATE Tokenize \(x\) into sentences \(\{s_j\}\)
\STATE In batch, for each \(j\): compute \(p_j \leftarrow f(x \text{ without } s_j)[y]\); \(q \leftarrow q + |\{s_j\}|\)
\STATE Pick \(s^{*} \leftarrow \arg\min_j p_j\) \COMMENT{largest drop}
\STATE \(N \leftarrow\) \textbf{if} \(m=\text{auto}\) \textbf{then} \(\mathrm{DynN}(|s^{*}|,ds)\) \textbf{else} \(N_{man}\)
\STATE Initialize \(\mathcal{T}\) rooted at span covering \(s^{*}\)
\WHILE{true}
  \STATE \((i, \Delta q, \mathcal{T}) \leftarrow \mathrm{N\_Nary\_Select\_Iter}(f, x^{cur}, y, N, \mathcal{T})\); \(q \leftarrow q + \Delta q\)
  \STATE \((s, x^{cur}, p, \Delta q) \leftarrow \mathrm{WordNetReplace}(f, x^{cur}, y, p_0, i)\); \(q \leftarrow q + \Delta q\); \(k \leftarrow k - 1\)
  \IF{\(s=\textbf{true}\)} \STATE \textbf{return} (\textbf{true}, \(x^{cur}\), \(q\), \(p\)) \ENDIF
  \IF{\(k=0\) \textbf{ or } no unexplored node in \(\mathcal{T}\)} \STATE \textbf{return} (\textbf{false}, \(x^{cur}\), \(q\), \(p_0\)) \ENDIF
\ENDWHILE
\end{algorithmic}
\end{minipage}
\end{figure*}

% ---------- HybridDynN_WNR (Token-level Hybrid) ----------
\begin{figure*}[t]
\centering
\begin{minipage}{0.97\textwidth}
\captionof{algorithm}{HybridDynN\_WNR (Token-level Hybrid)}
\label{alg:hyb-dynn}
\begin{algorithmic}[1]
\STATE \textbf{Input:} \(f, x, y, ds, m, N_{man}, k, \texttt{replace}\)
\STATE \textbf{Output:} success flag, \(x^{cur}\), \(q\), final prob \(p\)
\STATE \(N \leftarrow\) \textbf{if} \(m=\text{auto}\) \textbf{then} \(\mathrm{DynN}(|x|,ds)\) \textbf{else} \(N_{man}\)
\STATE \(p_0 \leftarrow f(x)[y]\); \(q \leftarrow 1\); \(x^{cur} \leftarrow x\); \(\mathcal{T} \leftarrow \emptyset\)
\WHILE{true}
  \STATE \((i, \Delta q, \mathcal{T}) \leftarrow \mathrm{N\_Nary\_Select\_Iter}(f, x^{cur}, y, N, \mathcal{T})\); \(q \leftarrow q + \Delta q\)
  \STATE \COMMENT{(Optional) extra greedy scoring around \(i\) using batched token removals to refine choice}
  \STATE \((s, x^{cur}, p, \Delta q) \leftarrow \mathrm{WordNetReplace}(f, x^{cur}, y, p_0, i)\); \(q \leftarrow q + \Delta q\); \(k \leftarrow k - 1\)
  \IF{\(s=\textbf{true}\)} \STATE \textbf{return} (\textbf{true}, \(x^{cur}\), \(q\), \(p\)) \ENDIF
  \IF{\(k=0\) \textbf{ or } no unexplored node in \(\mathcal{T}\)} \STATE \textbf{return} (\textbf{false}, \(x^{cur}\), \(q\), \(p_0\)) \ENDIF
\ENDWHILE
\end{algorithmic}
\end{minipage}
\end{figure*}

% ---------- HybridPure_WNR (No Sentence Stage / Pure Hybrid) ----------
\begin{figure*}[t]
\centering
\begin{minipage}{0.97\textwidth}
\captionof{algorithm}{HybridPure\_WNR (No Sentence Stage / Pure Hybrid)}
\label{alg:hybrid-pure}
\begin{algorithmic}[1]
\STATE \textbf{Input:} \(f, x, y, ds, m, N_{man}, k, \texttt{replace}, \tau\) \COMMENT{\(\tau\)=threshold fraction}
\STATE \textbf{Output:} success flag, \(x^{cur}\), \(q\), final prob \(p\)
\STATE \(N \leftarrow\) \textbf{if} \(m=\text{auto}\) \textbf{then} \(\mathrm{DynN}(|x|,ds)\) \textbf{else} \(N_{man}\)
\STATE \(p_0 \leftarrow f(x)[y]\); \(q \leftarrow 1\); \(x^{cur} \leftarrow x\); \(\mathcal{T} \leftarrow \emptyset\)
\WHILE{true}
  \STATE \(([a,b], \Delta q, \mathcal{T}) \leftarrow \mathrm{N\_Nary\_Select\_Segment}(f, x^{cur}, y, N, \mathcal{T}, \tau)\); \(q \leftarrow q + \Delta q\)
  \IF{\([a,b]=\emptyset\)} \STATE \textbf{return} (\textbf{false}, \(x^{cur}\), \(q\), \(p_0\)) \ENDIF
  \STATE \COMMENT{Greedy per-token scoring within chosen segment (batched removal)}
  \STATE For each \(t \in [a,b]\): form \(x^{(-t)}\) by dropping token \(t\); compute \(p_t \leftarrow f(x^{(-t)})[y]\)
  \STATE \(q \leftarrow q + (b-a+1)\); \(i \leftarrow \arg\min_{t \in [a,b]} p_t\)
  \IF{\(\texttt{replace}=\text{wordnet}\)}
    \STATE \((s, x^{cur}, p, \Delta q) \leftarrow \mathrm{WordNetReplace}(f, x^{cur}, y, p_0, i)\)
  \ELSE
    \STATE \((s, x^{cur}, p, \Delta q) \leftarrow \mathrm{BERTReplace}(f, x^{cur}, y, p_0, i)\)
  \ENDIF
  \STATE \(q \leftarrow q + \Delta q\); \(k \leftarrow k - 1\)
  \IF{\(s=\textbf{true}\)} \STATE \textbf{return} (\textbf{true}, \(x^{cur}\), \(q\), \(p\)) \ENDIF
  \IF{\(k=0\) \textbf{ or } no unexplored node in \(\mathcal{T}\)} \STATE \textbf{return} (\textbf{false}, \(x^{cur}\), \(q\), \(p_0\)) \ENDIF
\ENDWHILE
\end{algorithmic}
\end{minipage}
\end{figure*}

\newpage
\onecolumn
\twocolumn
\section{Related Work}

Adversarial text attacks have emerged as a critical area of research for testing the robustness of Natural Language Processing (NLP) models against manipulative inputs. These attacks, spanning various levels such as character, word, phrase, sentence, and multi-level manipulations, aim to exploit vulnerabilities in NLP systems. Early work, such as HotFlip \citep{ebrahimi-etal-2018-hotflip}, introduced white-box attacks by leveraging gradient-based strategies to replace words and deceive classifiers. White-box attacks, which have complete access to model information, including weights and architecture \cite{sadrizadeh-2022-block, wang-etal-2022-semattack}, are efficient in identifying impactful perturbations due to their comprehensive model knowledge. However, black-box attacks have gained prominence for their practicality in real-world scenarios where only confidence levels or predicted labels are accessible \cite{le-etal-2022-perturbations,jin2020bert}. Foundational work, such as \textit{Generating Natural Language Adversarial Examples} \citep{alzantot-etal-2018-generating}, pioneered black-box techniques by utilizing word saliency scores to craft adversarial examples, demonstrating the feasibility of attacking models without direct access to their internals.

At the character level, adversarial methods manipulate individual characters to disrupt tokenization processes. Techniques include adding or removing whitespace \cite{Grndahl2018AllYN}, replacing characters with visually similar alternatives \cite{eger-etal-2019-text}, or shuffling characters \cite{Li2019Textbugger}. Word-level attacks focus on synonym replacements, leveraging resources like Word Embeddings \cite{hsieh-etal-2019-robustness}, WordNet \cite{ren-etal-2019-generating}, and Mask Language Models \cite{li-etal-2020-bert-attack} to identify substitutes that maintain semantic coherence while deceiving classifiers. Phrase-level approaches replace multiple consecutive words \cite{deng-etal-2022-valcat, lei-etal-2022-phrase}, while sentence-level techniques generate adversarial text to exploit model weaknesses \cite{ribeiro-etal-2018-semantically, zhao2018generating}. Multi-level attacks combine these methods for more comprehensive perturbations \cite{formento-etal-2023-using}. Though our research focuses on word-level attacks, the methodology can extend to character or phrase-level attacks seamlessly.

Transformer-based models like BERT have revealed new vulnerabilities in NLP systems. For example, \textit{BAE: BERT-Based Adversarial Examples for Text Classification} \citep{garg-ramakrishnan-2020-bae} and \textit{BERT-ATTACK} \citep{li-etal-2020-bertattack} highlight how contextual embeddings can be exploited. These attacks underscore the need for enhanced defenses, particularly as adversarial attacks evolve. Defensive strategies like adversarial training have emerged as a popular countermeasure, incorporating adversarial examples during training to bolster robustness. For instance, \textit{Natural Language Adversarial Defense through Synonym Encoding} \citep{wang-etal-2020-synonym} and \textit{Towards Improving Adversarial Training of NLP Models} \citep{zhu-etal-2021-adversarial-training} demonstrate promising advancements. However, challenges persist in achieving generalizable defenses that can withstand novel attacks.

To further address robustness, behavioral testing frameworks such as \textit{Beyond Accuracy: Behavioral Testing of NLP Models with CheckList} \citep{ribeiro-etal-2020-checklist} have been introduced. These methodologies systematically evaluate model performance under diverse adversarial scenarios, emphasizing the importance of comprehensive testing to uncover potential weaknesses. Nevertheless, achieving a balance between attack success rates, query efficiency, and interpretability remains a significant challenge. Research like \textit{Pathologies of Neural Models Make Interpretations Difficult} \citep{saphra-lopez-2018-pathologies} highlights the difficulty of interpreting adversarial perturbations, while \textit{Certified Robustness to Adversarial Word Substitutions} \citep{jia-etal-2019-certified} explores trade-offs between computational efficiency and robustness guarantees.

\section{Comparison with Previous Greedy Attacks}\label{sect:baselines}
We additionally compare our proposed selection methods with two other attacks which leverage a variation of GreedySelect: PWWS \cite{ren-etal-2019-generating} and CLARE \cite{li2021contextualized}. The results can be found in Table \ref{tab:comparison_results}. As can be observed, both PWWS and CLARE use a significantly higher amount of queries. For example on IMDB, PWWS uses over 4 times as many queries as our most efficient method (Sentence Level Hybrid) and CLARE uses 81 times as many. CLAREs higher use of queries allow for a higher ASR and Similarity and PWWS achieves higher similarity and lower perturbation rate on average. This highlights another tradeoff between efficiency and subtlety of attack, as our methods require 2 to 4 times more perturbation rates. 

% Table: Rotten Tomatoes Dataset
\begin{table*}[!h]
\resizebox{\textwidth}{!}{%
\renewcommand{\arraystretch}{1.5}
\footnotesize
\begin{tabular}{c|c|c|c|c|c|c||c|c|c|c|}
\hline
 
 & & & \multicolumn{4}{|c|}{\textbf{DistilBert} } & \multicolumn{4}{|c|}{\textbf{DeBERTa}} \\\hline
 
& \textbf{Method} & \textbf{$N$} & \textbf{ASR}  & \textbf{AvgQ} & \textbf{Sim.} & \textbf{Pert.} & \textbf{ASR} & \textbf{AvgQ} & \textbf{Sim.} & \textbf{Pert.}\\ \hline

% IMDB
 \parbox[t]{2mm}{\multirow{6}{*}{\rotatebox[origin=c]{90}{IMDB}}} & PWWS  & - & 100 & 1498  & 96 & 5  & 56 & 1391 & 96  & 4\\ \cline{2-2} \cline{3-11}
 & CLARE  & - & 100  & 29460   & 98  & 1  & 51* & 22637 & 98& 1\\ \cline{2-2} \cline{3-11} 

 & Hybrid & 3 & 98  & 392   & 91  & 8& 70 &302 & 91& 8\\  \cline{3-11}

 & Sent. Hybrid & 3 & 98  & 362  &  90 & 9& 71 & 302 & 90&9\\  \cline{3-11}

 & Dynamic & - & 98  & 456& 90 & 8&70 &384 & 82& 19\\  \cline{3-11}
 & Dynamic  + Hybrid & - & 99 & 395  & 90  & 8& 71& 312& 82& 19\\ \hline\hline

% Yelp
 \parbox[t]{2mm}{\multirow{6}{*}{\rotatebox[origin=c]{90}{Yelp}}} & PWWS  & - & 99  & 934  & 94  &  6 & 97  & 985 & 92  & 8 \\ \cline{2-2} \cline{3-11}
 & CLARE  & - &  100 & 20194  & 97  & 3  & 99 & 22107& 96& 3\\ \cline{2-2} \cline{3-11} 

 & Hybrid &  2 & 98  & 363  & 87&12&95 &472 &81&18\\  \cline{3-11}

 & Sent. Hybrid & 2 & 97  &327   &87& 12&95 & 403& 81&18\\  \cline{3-11}

 & Dynamic & - &  97 & 378& 86& 12&96 & 525& 80& 18\\  \cline{3-11}
 & Dynamic  + Hybrid & - & 97 & 319  & 86& 12& 96 & 462 & 80& 18\\ \hline\hline

% AG News

 \parbox[t]{2mm}{\multirow{6}{*}{\rotatebox[origin=c]{90}{AG News}}} & PWWS  & - & 62  & 338  & 88 & 20  & 63 & 333  & 89 & 22\\ \cline{2-2} \cline{3-11}
 & CLARE  & - & 97  & 8113  & 91  & 8  & 92 &8541 &91 & 8\\ \cline{2-2} \cline{3-11} 

 & Hybrid & 3 & 72  &138   & 74  & 28 &77 &138 & 73&30\\  \cline{3-11}

 & Sent. Hybrid & 3 & 73  & 164  & 74  & 28 & 77&162 &73 &28\\  \cline{3-11}

 & Dynamic & - & 72  &169 & 73 & 32 &77 & 167& 73 & 22 \\  \cline{3-11}
 & Dynamic  + Hybrid & - & 72 & 141  & 73  & 32  & 77 & 140 & 74 & 33  \\ \hline\hline

 % Rotten Tomatoes
 \parbox[t]{2mm}{\multirow{6}{*}{\rotatebox[origin=c]{90}{Rotten Tomatoes}}} & PWWS  & - & 86 & 139 & 86 & 12   & 81 & 147 & 84 & 13 \\ \cline{2-2} \cline{3-11}
 & CLARE  & - & 100  & 1605  &  88 & 7  & 100 & 1668& 87& 7\\ \cline{2-2} \cline{3-11} 

 & Hybrid & 2 & 78  & 38  & 80& 15& 89& 49&75&19\\  \cline{3-11}

 & Sent. Hybrid & 2 & 78  & 38  & 79& 16& 90& 65& 75&19\\  \cline{3-11}

 & Dynamic & - & 78  & 50& 82& 14& 91& 63& 76& 20\\  \cline{3-11}
 & Dynamic  + Hybrid & - & 78 & 37  & 82& 14& 91 & 63 & 76& 20\\ \hline\hline
\end{tabular}%
}

\caption{Results for different methods on the noted datasets where ASR is attack success rate, AvgQ is the average queries by the successful attacks, Sim. is the similarity of text to the original, and Pert. is the percentage of text modified. A $t$ of 10\% was used for all Hybrid methods. \textbf{Bold} values indicate the best for that column, \underline{underline} indicates second best. * - Clare - IMDB was limited to 25000 query budget due to hardware constraints.}
\label{tab:comparison_results}
\end{table*}

\section{Ablation Studies} \label{sect:appendix}

To better understand the individual contributions of each component in our proposed attack strategies, we performed a series of ablation experiments. These experiments isolate the effects of selection algorithms. In the following, we provide a detailed analysis of the results.

\subsection{Effect of $N$ in N-Nary Algorithm}\label{sect:ablationN}

The N-Nary Select approach introduces hierarchical splitting, and we assess the impact of varying \(n\) (e.g., 3-Nary, 6-Nary, 12-Nary, 24-Nary). Lower values of \(n\) balance efficiency and effectiveness, achieving high attack success rates (ASR) with moderate query counts. For instance, on the IMDB dataset, 3-Nary Select achieves an ASR of 98\% with an average query count of 475, while 24-Nary achieves a similar ASR with 747 queries. This highlights a trade-off between finer granularity and computational overhead.

On the AGNews dataset, where text length is shorter, higher \(n\) values show diminishing returns. For example, 6-Nary achieves no significant improvement in ASR over 3-Nary but increases the average query count by over 17\%. These results underscore the importance of selecting an appropriate \(n\) based on dataset characteristics.

%\subsubsection{Algorithm Performance Analysis}

Among the algorithms evaluated, the 3-nary Select consistently achieves the best balance between query efficiency and attack accuracy. It performs  well at lower split thresholds, with fewer queries required to achieve successful attacks compared to Binary and 6-nary searches.

\begin{figure}[t]
    \centering
    \includegraphics[width=\columnwidth,height=1.0\textheight,keepaspectratio]{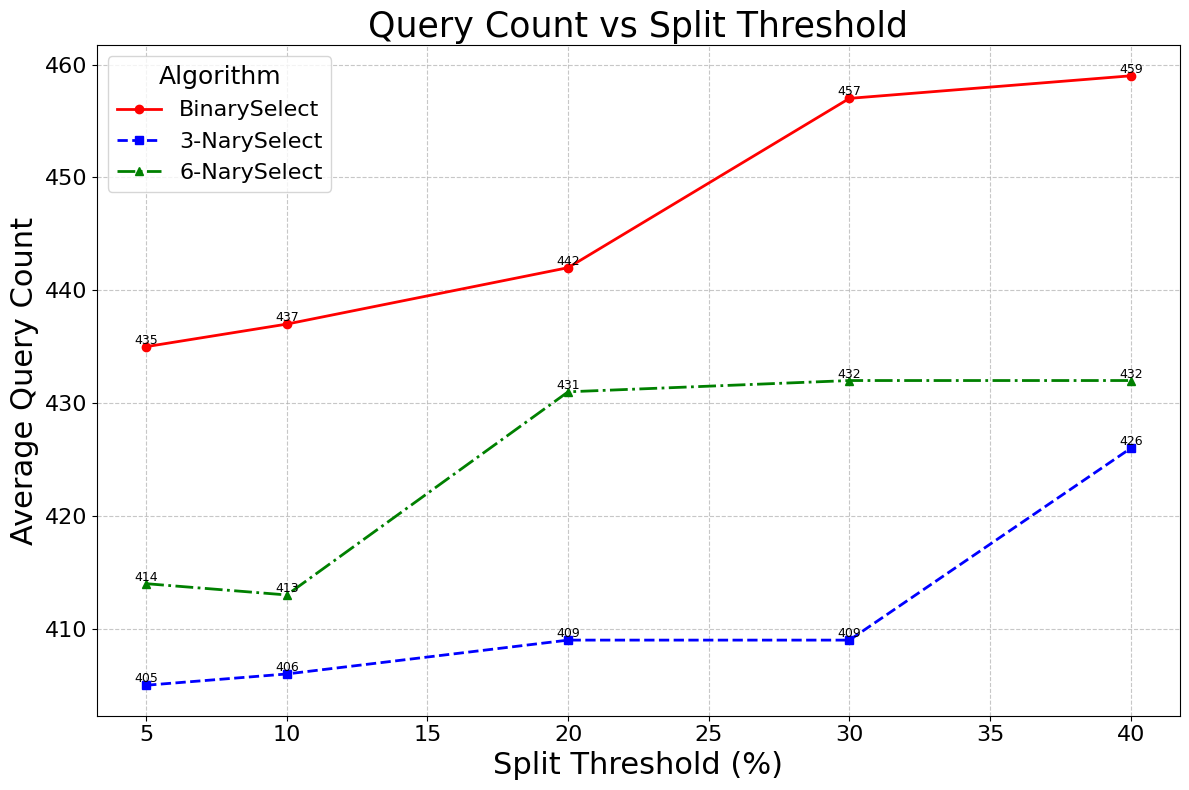}
    \caption{Query count versus Split Threshold $t$ for Hybrid N-nary. }
    \label{fig:Threshold_QryCnt}
\end{figure}

\begin{figure}[t]
    \centering
    \includegraphics[width=\columnwidth,height=2.0\textheight,keepaspectratio]{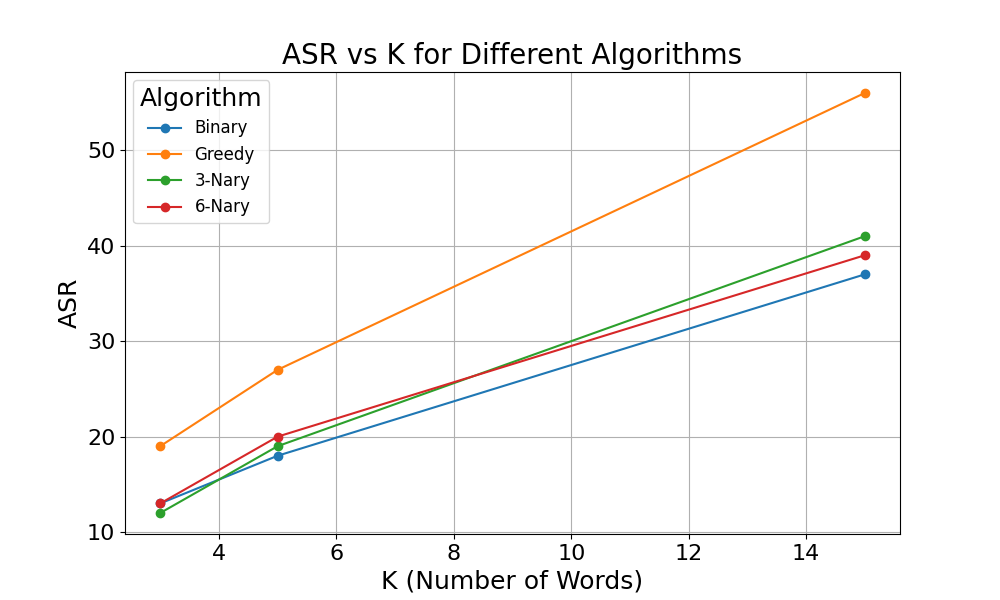}
    \caption{$k$ (number of words allowed to be modified) versus ASR for base N-nary Select Algorithms. Results are consistent with previous BinarySelect work.}
    \label{fig:k_ASR_Nnary}
\end{figure}

\subsection{Impact of Split Threshold $t$ for Hybrid N-nary}\label{sect:ablationthreshold}

We analyze the impact of varying split thresholds on the performance of the hybrid Select algorithms (Binary, 3-nary, and 6-nary) as shown in Figure \ref{fig:Threshold_QryCnt}. 

As shown in the Figure \ref{fig:Threshold_QryCnt}, lower split thresholds, such as 5\% and 10\%, consistently result in reduced average query counts for successful attacks. This is particularly evident in the 3-nary algorithm, which demonstrates the highest query efficiency across most thresholds. The finer-grained splits enabled by lower thresholds allow for more precise identification of impactful word substitutions, minimizing the number of queries required.

In contrast, higher split thresholds, such as 30\% and 40\%, generate larger segments that demand more exploration to identify optimal substitutions. While this approach captures more context, it significantly increases the query count, making it less efficient compared to finer splits.

Among the algorithms, the 6-nary algorithm exhibits slightly higher query counts at lower thresholds but maintains comparable attack accuracy. Meanwhile, the Binary Select consistently performs the least efficiently across all thresholds, requiring the highest query counts due to its limited segmentation and exploration capabilities.

\subsection{Word Replacement Strategies}\label{sect:wordreplacement}
Alongside Wordnet replacement, we also explored a BERT focused replacement method by using mask-infill method where we mask the most influential word and find a replacement candidate (up to 5) using BERT. Table \ref{tab:imdb_bert} (Appendix \ref{sect:imdb_table}) shows the results for N-Nary attack in IMDB using BERT replacement. From the results, there are no significant improvements when compared to the Wordnet replacement method.

\subsection{The Impact of \( k \) on Query Count and Attack Success Rate (ASR)}\label{sect:ablationk}

We also evaluate how varying \( k \) (number of words allowed for replacement) affects query count and attack success rate (ASR) across standalone n-nary search methods (Binary, 3-Nary, 6-Nary, and Greedy) and the hybrid method with segmentation thresholds. Both \( K \) and segmentation thresholds present a trade-off between query efficiency and ASR. Figure \ref{fig:k_ASR_Nnary} shows how ASR changes as $k$ changes.

%\subsubsection{Impact of \( K \) on Query Count and ASR}

In both standalone and hybrid methods, smaller \( k \) values (e.g., \( k=3 \)) reduce query counts as fewer replacements are allowed, narrowing the search space. For instance, at \( k=3 \), the 3-Nary method achieves the lowest query count. However, smaller \( k \) values limit ASR, as the restricted replacement budget fails to significantly degrade the model’s accuracy. Across methods, ASR at \( k=3 \) remains modest.

As \( k \) increases, query counts rise proportionally across all methods. At \( k=15 \), ASR improves significantly as more words are replaced, allowing for more impactful attacks. The Greedy algorithm achieves the highest ASR at this configuration, though it requires more queries. Among n-nary methods, 3-Nary consistently strikes the best balance between query efficiency and ASR, while 6-Nary incurs the highest query counts due to its complex segmentation.

\subsection{Dynamic-N Bins}
Table \ref{tab:optimal-n} (Appendix \ref{DynamicN_bins}) contains the length bins and their corresponding Optimal-N values for IMDB, Yelp, and AGNews dataset. Based on the maximum length of the input text in validation dataset, 5 length bins are created and their corresponding optimal N values are obtained based on the query count. From the table, the n-value of \textit{3} seems to be more optimal for most text length bins for the IMDB and Yelp datasets, \textit{2} for AGNews. This also matches the results for hybrid and standalone methods, as the n-value of \textit{3} has been shown to perform better in most cases.
%what diff bins mean
% \input{RT_distilbert_table}
\newpage
\onecolumn
\section{Length Bins and their Optimal-N}\label{DynamicN_bins}
\begin{table*}[h!]
    \resizebox{\textwidth}{!}{% Reduce the size of the table to fit the text width
    \renewcommand{\arraystretch}{0.5}
    \begin{tabular}{ccc}
        % Table 1
        \begin{minipage}{0.32\textwidth}
            \centering
            \begin{tabular}{cc}
            \toprule
            \textbf{Length Bin}  & \textbf{n}  \\
            \midrule
            {1 - 200} & 3   \\
            \midrule
            {200 - 400}  & 3   \\
            \midrule
            {400 - 600}  & 3  \\
            \midrule
            {600 - 800}  & 2  \\
            \midrule
            {800 - max} & 6  \\
            \bottomrule
        \end{tabular}
            \caption{IMDB}
            \label{tab:query-count-1}
        \end{minipage} &
        % Table 2
        \begin{minipage}{0.32\textwidth}
            \centering
            \begin{tabular}{cc}
            \toprule
            \textbf{Length Bin}  & \textbf{n}  \\
            \midrule
            {1 - 135} & 3   \\
            \midrule
            {135 - 270}  & 3   \\
            \midrule
            {270 - 405}  & 3  \\
            \midrule
            {405 - 540}  & 2  \\
            \midrule
            {540 - max} & 6  \\
            \bottomrule
        \end{tabular}
            \caption{Yelp}
            \label{tab:query-count-2}
        \end{minipage} &
        % Table 3
        \begin{minipage}{0.32\textwidth}
            \centering
            \begin{tabular}{cc}
            \toprule
            \textbf{Length Bin}  & \textbf{n}  \\
            \midrule
            {1 - 23} & 2   \\
            \midrule
            {23 - 46}  & 2   \\
            \midrule
            {46 - 69}  & 2  \\
            \midrule
            {69 - 92}  & 2  \\
            \midrule
            {92 - max} & 2  \\
            \bottomrule
        \end{tabular}
            \caption{AGNews}
            \label{tab:query-count-3}
        \end{minipage}
    \end{tabular}
    }
    \caption{Length bins and their Optimal-N}
    \label{tab:optimal-n}
\end{table*}
\section{IMDB Tables} \label{sect:imdb_table}
\begin{table*}[ht]
\centering
\caption{Results Summary for N-nary Algorithm and Baseline(Binary and Greedy) with WordNet Replacement Strategy}
\renewcommand{\arraystretch}{1.5}
\resizebox{\textwidth}{!}{%
\begin{tabular}{|c|c|c|c|c|c|c|}
\hline
\textbf{K (No. of Words)} & \textbf{Algorithm} & \textbf{Original Accuracy (\%)} & \textbf{Avg Query} & \textbf{Avg Query (For Atk Success)} & \textbf{Attack Accuracy (\%)} & \textbf{ASR (\%)} \\\hline
ALL & Binary & 92.0 & 504 & 484 & 1.7 & 98 \\
    & Greedy & 92.0 & 433 & 425 & 1.2 & 99 \\
    & 3-Nary & 92.0 & 475 & 461 & 1.5 & 98 \\
    & 6-Nary & 92.0 & 585 & 569 & 1.3 & 99 \\\hline
3   & Binary & 92.0 & 48 & 32 & 80.4 & 13 \\
    & Greedy & 92.0 & 238 & 215 & 75.0 & 18 \\
    & 3-Nary & 92.0 & 46 & 35 & 81.1 & 12 \\
    & 6-Nary & 92.0 & 52 & 38 & 80.1 & 13 \\\hline
5   & Binary & 92.0 & 66 & 42 & 75.9 & 17 \\
    & Greedy & 92.0 & 247 & 216 & 67.5 & 27 \\
    & 3-Nary & 92.0 & 64 & 45 & 74.3 & 19 \\
    & 6-Nary & 92.0 & 71 & 50 & 73.9 & 20 \\\hline
15  & Binary & 92.0 & 141 & 82 & 58.2 & 37 \\
    & Greedy & 92.0 & 279 & 250 & 40.9 & 56 \\
    & 3-Nary & 92.0 & 133 & 83 & 54.5 & 41 \\
    & 6-Nary & 92.0 & 139 & 90 & 56.2 & 39 \\\hline
\end{tabular}%
}
\label{tab:results_summary}
\end{table*}

\begin{table*}[ht]
\centering
\caption{Results Summary for N-Nary and Greedy Algorithms using BERT replacement stategy}
\renewcommand{\arraystretch}{1.5}
\resizebox{\textwidth}{!}{%
\begin{tabular}{|c|c|c|c|c|c|c|}
\hline
\textbf{K (No. of Words)} & \textbf{Algorithm} & \textbf{Original Accuracy (\%)} & \textbf{Avg Query} & \textbf{Avg Query (For Atk Success)} & \textbf{Attack Accuracy (\%)} & \textbf{ASR (\%)} \\\hline
ALL & Binary & 92 & 504 & 483 & 1.8 & 98 \\
    & Greedy & 92 & 432 & 422 & 1.4 & 99 \\
    & 3-Nary & 92 & 473 & 458 & 1.5 & 98 \\
    & 6-Nary & 92 & 582 & 570 & 1.3 & 99 \\\hline
3   & Binary & 92 & 47 & 31 & 80.4 & 13 \\
    & Greedy & 92 & 238 & 214 & 75.0 & 18 \\
    & 3-Nary & 92 & 46 & 33 & 81.0 & 12 \\
    & 6-Nary & 92 & 52 & 37 & 80.1 & 13 \\\hline
5   & Binary & 92 & 66 & 41 & 75.9 & 17 \\
    & Greedy & 92 & 247 & 215 & 67.5 & 26 \\
    & 3-Nary & 92 & 64 & 44 & 74.3 & 19 \\
    & 6-Nary & 92 & 71 & 50 & 73.9 & 20 \\\hline
15  & Binary & 92 & 141 & 82 & 58.2 & 37 \\
    & Greedy & 92 & 279 & 249 & 40.9 & 55 \\
    & 3-Nary & 92 & 133 & 83 & 54.4 & 41 \\
    & 6-Nary & 92 & 138 & 89 & 56.2 & 39 \\\hline
\end{tabular}%
}
\label{tab:imdb_bert}
\end{table*}

% Table for 5% Split Threshold
\begin{table*}[ht]
\centering
\caption{Results Summary for Hybrid Algorithm at Sentence Level (5\% Split Threshold) with WordNet Replacement Strategy}
\renewcommand{\arraystretch}{1.5}
\resizebox{\textwidth}{!}{%
\begin{tabular}{|c|c|c|c|c|c|c|c|}
\hline
\textbf{K (No. of Words)} & \textbf{Algorithm} & \textbf{Original Accuracy (\%)} & \textbf{Avg Query} & \textbf{Avg Query (For Atk Success)} & \textbf{Avg Query (For Fail)} & \textbf{Attack Accuracy (\%)} & \textbf{ASR (\%)} \\\hline
ALL & Binary & 92 & 375 & 363 & 998 & 1.8 & 98 \\
    & 3-Nary & 92 & 364 & 1053 & 1053 & 1.6 & 98 \\
    & 6-Nary & 92 & 392 & 380 & 974 & 1.9 & 98 \\\hline
3   & Binary & 92 & 51 & 37 & 53 & 79.2 & 14 \\
    & 3-Nary & 92 & 49 & 32 & 52 & 80.3 & 13 \\
    & 6-Nary & 92 & 50 & 34 & 52 & 79.9 & 13 \\\hline
5   & Binary & 92 & 66 & 44 & 71 & 74.6 & 17 \\
    & 3-Nary & 92 & 65 & 43 & 71 & 74.2 & 18 \\
    & 6-Nary & 92 & 66 & 43 & 71 & 74.8 & 17 \\\hline
15  & Binary & 92 & 126 & 78 & 157 & 55.6 & 40 \\
    & 3-Nary & 92 & 124 & 79 & 156 & 54.3 & 41 \\
    & 6-Nary & 92 & 126 & 83 & 156 & 54.2 & 41 \\\hline
\end{tabular}%
}
\label{tab:5_percent_threshold}
\end{table*}

% Table for 10% Split Threshold
\begin{table*}[ht]
\centering
\caption{Results Summary for Hybrid Algorithm at Sentence Level (10\% Split Threshold) with WordNet Replacement Strategy}
\renewcommand{\arraystretch}{1.5}
\resizebox{\textwidth}{!}{%
\begin{tabular}{|c|c|c|c|c|c|c|c|}
\hline
\textbf{K (No. of Words)} & \textbf{Algorithm} & \textbf{Original Accuracy (\%)} & \textbf{Avg Query} & \textbf{Avg Query (For Atk Success)} & \textbf{Avg Query (For Fail)} & \textbf{Attack Accuracy (\%)} & \textbf{ASR (\%)} \\\hline
ALL & Binary & 92 & 378 & 368 & 943 & 1.7 & 98 \\
    & 3-Nary & 92 & 371 & 363 & 865 & 1.5 & 98 \\
    & 6-Nary & 92 & 387 & 374 & 945 & 2.1 & 98 \\\hline
3   & Binary & 92 & 54 & 38 & 56 & 79.9 & 14 \\
    & 3-Nary & 92 & 51 & 34 & 54 & 80.2 & 13 \\
    & 6-Nary & 92 & 50 & 34 & 52 & 79.6 & 14 \\\hline
5   & Binary & 92 & 70 & 46 & 75 & 75.1 & 17 \\
    & 3-Nary & 92 & 68 & 43 & 73 & 74.5 & 18 \\
    & 6-Nary & 92 & 66 & 44 & 71 & 74.3 & 18 \\\hline
15  & Binary & 92 & 129 & 81 & 160 & 55.5 & 40 \\
    & 3-Nary & 92 & 128 & 81 & 159 & 54.8 & 41 \\
    & 6-Nary & 92 & 125 & 81 & 156 & 54.1 & 41 \\\hline
\end{tabular}%
}
\label{tab:10_percent_threshold}
\end{table*}
\begin{table*}[ht]
\centering
\caption{Results Summary for Hybrid N-Nary 5\% Split Threshold with WordNet Replacement Strategy}
\renewcommand{\arraystretch}{1.5}
\resizebox{\textwidth}{!}{%
\begin{tabular}{|c|c|c|c|c|c|c|c|}
\hline
\textbf{K (No. of Words)} & \textbf{Algorithm} & \textbf{Original Accuracy (\%)} & \textbf{Avg Query} & \textbf{Avg Query (For Atk Success)} & \textbf{Avg Query (For Fail)} & \textbf{Attack Accuracy (\%)} & \textbf{ASR (\%)} \\\hline
ALL & Binary & 92 & 435 & 425 & 1251 & 1.2 & 99 \\
    & 3-Nary & 92 & 405 & 388 & 1363 & 1.6 & 98 \\
    & 6-Nary & 92 & 414 & 403 & 1200 & 1.3 & 98 \\\hline
3   & Binary & 92 & 49 & 35 & 51 & 80.3 & 13 \\
    & 3-Nary & 92 & 49 & 37 & 51 & 80.2 & 13 \\
    & 6-Nary & 92 & 50 & 35 & 52 & 80.5 & 13 \\\hline
5   & Binary & 92 & 65 & 43 & 70 & 75.5 & 17 \\
    & 3-Nary & 92 & 65 & 44 & 70 & 75.2 & 18 \\
    & 6-Nary & 92 & 65 & 43 & 70 & 75.0 & 18 \\\hline
15  & Binary & 92 & 127 & 80 & 155 & 57.5 & 38 \\
    & 3-Nary & 92 & 124 & 78 & 154 & 56.2 & 39 \\
    & 6-Nary & 92 & 123 & 79 & 152 & 55.3 & 40 \\\hline
\end{tabular}%
}
\label{tab:5_split_threshold}
\end{table*}

\begin{table*}[ht]
\centering
\caption{Results Summary for Hybrid N-Nary 10\% Split Threshold with WordNet Replacement Strategy}
\renewcommand{\arraystretch}{1.5}
\resizebox{\textwidth}{!}{%
\begin{tabular}{|c|c|c|c|c|c|c|c|}
\hline
\textbf{K (No. of Words)} & \textbf{Algorithm} & \textbf{Original Accuracy (\%)} & \textbf{Avg Query} & \textbf{Avg Query (For Atk Success)} & \textbf{Avg Query (For Fail)} & \textbf{Attack Accuracy (\%)} & \textbf{ASR (\%)} \\\hline
ALL & Binary & 92 & 437 & 427 & 1117 & 1.3 & 99 \\
    & 3-Nary & 92 & 406 & 392 & 1320 & 1.4 & 98 \\
    & 6-Nary & 92 & 413 & 398 & 1327 & 1.5 & 98 \\\hline
3   & Binary & 92 & 57 & 38 & 60 & 80.9 & 12 \\
    & 3-Nary & 92 & 49 & 35 & 51 & 80.5 & 13 \\
    & 6-Nary & 92 & 49 & 33 & 52 & 80.5 & 13 \\\hline
5   & Binary & 92 & 73 & 47 & 78 & 75.3 & 18 \\
    & 3-Nary & 92 & 65 & 43 & 70 & 75.2 & 18 \\
    & 6-Nary & 92 & 65 & 43 & 70 & 74.9 & 18 \\\hline
15  & Binary & 92 & 133 & 83 & 163 & 58.4 & 37 \\
    & 3-Nary & 92 & 124 & 77 & 153 & 56.8 & 39 \\
    & 6-Nary & 92 & 123 & 79 & 152 & 55.3 & 40 \\\hline
\end{tabular}%
}
\label{tab:10_split_threshold}
\end{table*}

\begin{table*}[ht]
\centering
\caption{Results Summary for Hybrid N-Nary (20\% Split Threshold) with WordNet Replacement Strategy}
\renewcommand{\arraystretch}{1.5}
\resizebox{\textwidth}{!}{%
\begin{tabular}{|c|c|c|c|c|c|c|c|}
\hline
\textbf{K (No. of Words)} & \textbf{Algorithm} & \textbf{Original Accuracy (\%)} & \textbf{Avg Query} & \textbf{Avg Query (For Atk Success)} & \textbf{Avg Query (For Fail)} & \textbf{Attack Accuracy (\%)} & \textbf{ASR (\%)} \\\hline
ALL & Binary & 92 & 442 & 429 & 1526 & 1.1 & 99 \\
    & 3-Nary & 92 & 409 & 399 & 1279 & 1.1 & 99 \\
    & 6-Nary & 92 & 431 & 418 & 1302 & 1.3 & 98 \\\hline
3   & Binary & 92 & 74 & 49 & 77 & 80.6 & 13 \\
    & 3-Nary & 92 & 71 & 45 & 74 & 80.2 & 13 \\
    & 6-Nary & 92 & 88 & 66 & 92 & 79.1 & 14 \\\hline
5   & Binary & 92 & 89 & 56 & 97 & 75.0 & 18 \\
    & 3-Nary & 92 & 86 & 55 & 94 & 74.3 & 18 \\
    & 6-Nary & 92 & 106 & 75 & 114 & 72.8 & 21 \\\hline
15  & Binary & 92 & 145 & 91 & 177 & 57.1 & 38 \\
    & 3-Nary & 92 & 141 & 90 & 173 & 56.0 & 39 \\
    & 6-Nary & 92 & 160 & 114 & 192 & 54.2 & 41 \\\hline
\end{tabular}%
}
\label{tab:hybrid_nnary}
\end{table*}

\begin{table*}[ht]
\centering
\caption{Results Summary for  Hybrid N-Nary 30\% Split Threshold with WordNet Replacement Strategy}
\renewcommand{\arraystretch}{1.5}
\resizebox{\textwidth}{!}{%
\begin{tabular}{|c|c|c|c|c|c|c|c|}
\hline
\textbf{K (No. of Words)} & \textbf{Algorithm} & \textbf{Original Accuracy (\%)} & \textbf{Avg Query} & \textbf{Avg Query (For Atk Success)} & \textbf{Avg Query (For Fail)} & \textbf{Attack Accuracy (\%)} & \textbf{ASR (\%)} \\\hline
ALL & Binary & 92 & 457 & 448 & 1136 & 1.2 & 99 \\
    & 3-Nary & 92 & 409 & 395 & 1430 & 1.2 & 99 \\
    & 6-Nary & 92 & 432 & 420 & 1302 & 1.3 & 98 \\\hline
3   & Binary & 92 & 102 & 73 & 107 & 79.8 & 13 \\
    & 3-Nary & 92 & 71 & 45 & 74 & 80.2 & 13 \\
    & 6-Nary & 92 & 88 & 66 & 92 & 79.1 & 14 \\\hline
5   & Binary & 92 & 117 & 82 & 125 & 74.1 & 19 \\
    & 3-Nary & 92 & 86 & 55 & 94 & 74.3 & 18 \\
    & 6-Nary & 92 & 106 & 75 & 114 & 72.8 & 21 \\\hline
15  & Binary & 92 & 168 & 117 & 204 & 53.8 & 42 \\
    & 3-Nary & 92 & 141 & 90 & 173 & 56.1 & 39 \\
    & 6-Nary & 92 & 160 & 114 & 192 & 54.3 & 41 \\\hline
\end{tabular}%
}
\label{tab:30_split_threshold}
\end{table*}

\begin{table*}[ht]
\centering
\caption{Results Summary for  Hybrid N-Nary 40\% Split Threshold with WordNet Replacement Strategy}
\renewcommand{\arraystretch}{1.5} % Adjust row height
\resizebox{\textwidth}{!}{%
\begin{tabular}{|c|c|c|c|c|c|c|c|}
\hline
\textbf{K (No. of Words)} & \textbf{Algorithm} & \textbf{Original Accuracy (\%)} & \textbf{Avg Query} & \textbf{Avg Query (For Atk Success)} & \textbf{Avg Query (For Fail)} & \textbf{Attack Accuracy (\%)} & \textbf{ASR (\%)} \\\hline
ALL & Binary & 92 & 459 & 449 & 1167 & 1.3 & 99 \\
    & 3-Nary & 92 & 426 & 411 & 1324 & 1.5 & 98 \\
    & 6-Nary & 92 & 432 & 418 & 1479 & 1.2 & 99 \\\hline
3   & Binary & 92 & 102 & 72 & 107 & 79.8 & 13 \\
    & 3-Nary & 92 & 123 & 88 & 129 & 78.7 & 14 \\
    & 6-Nary & 92 & 88 & 66 & 92 & 79.1 & 14 \\\hline
5   & Binary & 92 & 117 & 82 & 125 & 74.0 & 19 \\
    & 3-Nary & 92 & 136 & 102 & 146 & 71.6 & 22 \\
    & 6-Nary & 92 & 106 & 75 & 114 & 72.7 & 21 \\\hline
15  & Binary & 92 & 168 & 117 & 204 & 53.9 & 42 \\
    & 3-Nary & 92 & 186 & 140 & 223 & 50.3 & 45 \\
    & 6-Nary & 92 & 160 & 114 & 192 & 54.4 & 41 \\\hline
\end{tabular}%
}
\label{tab:40_split_threshold}
\end{table*}

% 5% Hybrid Threshold Table
\begin{table*}[ht]
\centering
\caption{Results Summary for 5\% Hybrid Threshold (Dyn-N + Hybrid) with WordNet Replacement Strategy}
\renewcommand{\arraystretch}{1.5}
\resizebox{\textwidth}{!}{%
\begin{tabular}{|c|c|c|c|c|c|c|c|}
\hline
\textbf{K (No. of Words)} & \textbf{Algorithm} & \textbf{Original Accuracy (\%)} & \textbf{Avg Query} & \textbf{Avg Query (For Atk Success)} & \textbf{Attack Accuracy (\%)} & \textbf{ASR (\%)} \\\hline
ALL & DynN+Hyb & 92 & 408 & 390 & 1.6 & 98 \\
3   & DynN+Hyb & 92 & 22 & 15 & 81.0 & 12 \\
5   & DynN+Hyb & 92 & 35 & 23 & 74.5 & 19 \\
15  & DynN+Hyb & 92 & 90 & 52 & 54.4 & 41 \\\hline
\end{tabular}%
}
\label{tab:5_hybrid_threshold}
\end{table*}

% 10% Hybrid Threshold Table
\begin{table*}[ht]
\centering
\caption{Results Summary for 10\% Hybrid Threshold (Dyn-N + Hybrid) with WordNet Replacement Strategy}
\renewcommand{\arraystretch}{1.5}
\resizebox{\textwidth}{!}{%
\begin{tabular}{|c|c|c|c|c|c|c|c|}
\hline
\textbf{K (No. of Words)} & \textbf{Algorithm} & \textbf{Original Accuracy (\%)} & \textbf{Avg Query} & \textbf{Avg Query (For Atk Success)} & \textbf{Attack Accuracy (\%)} & \textbf{ASR (\%)} \\\hline
ALL & DynN+Hyb & 92 & 403 & 395 & 1.2 & 99 \\
3   & DynN+Hyb & 92 & 22 & 16 & 80.9 & 12 \\
5   & DynN+Hyb & 92 & 35 & 24 & 74.5 & 19 \\
15  & DynN+Hyb & 92 & 90 & 52 & 54.5 & 41 \\\hline
\end{tabular}%
}
\label{tab:10_hybrid_threshold}
\end{table*}

% 20% Hybrid Threshold Table
\begin{table*}[ht]
\centering
\caption{Results Summary for 20\% Hybrid Threshold (Dyn-N + Hybrid) with WordNet Replacement Strategy}
\renewcommand{\arraystretch}{1.5}
\resizebox{\textwidth}{!}{%
\begin{tabular}{|c|c|c|c|c|c|c|c|}
\hline
\textbf{K (No. of Words)} & \textbf{Algorithm} & \textbf{Original Accuracy (\%)} & \textbf{Avg Query} & \textbf{Avg Query (For Atk Success)} & \textbf{Attack Accuracy (\%)} & \textbf{ASR (\%)} \\\hline
ALL & DynN+Hyb & 92 & 406 & 391 & 1.5 & 98 \\
3   & DynN+Hyb & 92 & 22 & 16 & 81.0 & 12 \\
5   & DynN+Hyb & 92 & 35 & 23 & 74.5 & 19 \\
15  & DynN+Hyb & 92 & 90 & 52 & 54.6 & 41 \\\hline
\end{tabular}%
}
\label{tab:20_hybrid_threshold}
\end{table*}

% 30% Hybrid Threshold Table
\begin{table*}[ht]
\centering
\caption{Results Summary for 30\% Hybrid Threshold (Dyn-N + Hybrid) with WordNet Replacement Strategy}
\renewcommand{\arraystretch}{1.5}
\resizebox{\textwidth}{!}{%
\begin{tabular}{|c|c|c|c|c|c|c|c|}
\hline
\textbf{K (No. of Words)} & \textbf{Algorithm} & \textbf{Original Accuracy (\%)} & \textbf{Avg Query} & \textbf{Avg Query (For Atk Success)} & \textbf{Attack Accuracy (\%)} & \textbf{ASR (\%)} \\\hline
ALL & DynN+Hyb & 92 & 405 & 392 & 1.5 & 98 \\
3   & DynN+Hyb & 92 & 22 & 15 & 80.9 & 12 \\
5   & DynN+Hyb & 92 & 35 & 23 & 74.5 & 19 \\
15  & DynN+Hyb & 92 & 90 & 51 & 54.5 & 41 \\\hline
\end{tabular}%
}
\label{tab:30_hybrid_threshold}
\end{table*}

% 40% Hybrid Threshold Table
\begin{table*}[ht]
\centering
\caption{Results Summary for 40\% Hybrid Threshold (Dyn-N + Hybrid) with WordNet Replacement Strategy}
\renewcommand{\arraystretch}{1.5}
\resizebox{\textwidth}{!}{%
\begin{tabular}{|c|c|c|c|c|c|c|c|}
\hline
\textbf{K (No. of Words)} & \textbf{Algorithm} & \textbf{Original Accuracy (\%)} & \textbf{Avg Query} & \textbf{Avg Query (For Atk Success)} & \textbf{Attack Accuracy (\%)} & \textbf{ASR (\%)} \\\hline
ALL & DynN+Hyb & 92 & 404 & 388 & 1.3 & 99 \\
3   & DynN+Hyb & 92 & 22 & 16 & 80.9 & 12 \\
5   & DynN+Hyb & 92 & 35 & 24 & 74.5 & 19 \\
15  & DynN+Hyb & 92 & 90 & 52 & 54.6 & 41 \\\hline
\end{tabular}%
}
\label{tab:40_hybrid_threshold}
\end{table*}

% Dyn-N 
\begin{table}[h!]
\resizebox{\textwidth}{!}{%
\renewcommand{\arraystretch}{1.5}
\begin{tabular}{|c|c|c|c|c|c|c|}
\hline
K (No. of words) & Algorithm & Original Accuracy (\%) & ASR (\%) & Avg Query & Avg Query (For Atk Success) & Attack Accuracy (\%) \\
\hline
ALL & Binary & 92 & 90.5 & 471 & 456 & 1.5 \\
\hline
3 & Binary & 92 & 11 & 47 & 33 & 81 \\
\hline
5 & Binary & 92 & 17.5 & 65 & 45 & 74.5 \\
\hline
15 & Binary & 92 & 37.6 & 133 & 83 & 54.4 \\
\hline
\end{tabular}%
}
\caption{Results Summary for Dynamic-N}
\end{table}

\newpage
\onecolumn
\section{Yelp Tables}

\begin{table}[h!]
\resizebox{\textwidth}{!}{%
\renewcommand{\arraystretch}{1.5}
\begin{tabular}{|c|c|c|c|c|c|c|}
\hline
K (No. of words) & Algorithm & Original Accuracy (\%) & ASR (\%) & Avg Query & Avg Query (For Atk Success) & Attack Accuracy (\%) \\
\hline
ALL & Binary & 97 & 98 & 401 & 400 & 2 \\
& Greedy & 97 & 98 & 327 & 329 & 2 \\
& 3-Nary & 97 & 98 & 375 & 374 & 2 \\
& 6-Nary & 97 & 98 & 448 & 449 & 2 \\
\hline
3 & Binary & 97 & 11 & 46 & 32 & 86 \\
& Greedy & 97 & 13 & 155 & 103 & 84 \\
& 3-Nary & 97 & 11 & 43 & 32 & 86 \\
& 6-Nary & 97 & 11 & 49 & 35 & 86 \\
\hline
5 & Binary & 97 & 17 & 66 & 43 & 80 \\
& Greedy & 97 & 21 & 166 & 106 & 76 \\
& 3-Nary & 97 & 18 & 61 & 42 & 79 \\
& 6-Nary & 97 & 17 & 69 & 46 & 80 \\
\hline
15 & Binary & 97 & 41 & 139 & 87 & 57 \\
& Greedy & 97 & 52 & 207 & 155 & 45 \\
& 3-Nary & 97 & 42 & 133 & 86 & 55 \\
& 6-Nary & 97 & 39 & 138 & 88 & 58 \\
\hline
\end{tabular}%
}
\caption{Performance results for N-Nary in Yelp}
\end{table}

\begin{table}[h!]
\resizebox{\textwidth}{!}{%
\renewcommand{\arraystretch}{1.5}
\begin{tabular}{|c|c|c|c|c|c|c|c|}
\hline
\textbf{K (No. of words)} & \textbf{Replacement Strategy} & \textbf{Algorithm} & \textbf{Original Accuracy (\%)} & \textbf{ASR (\%)} & \textbf{Avg Query} & \textbf{Avg Query (For Atk Success)} & \textbf{Attack Accuracy (\%)} \\
\hline
\multirow{4}{*}{ALL} & \multirow{4}{*}{BERT (top\_k = 5)} 
    & Binary  & 97 & 94.6 & 403 & 402 & 2.4 \\
    & & Greedy  & 97 & 94.9 & 326 & 328 & 2.1 \\
    & & 3-Nary  & 97 & 94.6 & 375 & 374 & 2.4 \\
    & & 6-Nary  & 97 & 95.3 & 449 & 449 & 1.7 \\
\hline
\multirow{4}{*}{3} & \multirow{4}{*}{BERT (top\_k = 5)} 
    & Binary  & 97 & 11 & 46 & 33 & 86 \\
    & & Greedy  & 97 & 13.1 & 155 & 103 & 83.9 \\
    & & 3-Nary  & 97 & 11.4 & 43 & 32 & 85.6 \\
    & & 6-Nary  & 97 & 11.2 & 50 & 36 & 85.8 \\
\hline
\multirow{4}{*}{5} & \multirow{4}{*}{BERT (top\_k = 5)} 
    & Binary  & 97 & 16.9 & 66 & 44 & 80.1 \\
    & & Greedy  & 97 & 21.4 & 166 & 106 & 75.6 \\
    & & 3-Nary  & 97 & 17.6 & 61 & 41 & 79.4 \\
    & & 6-Nary  & 97 & 16.6 & 69 & 47 & 80.4 \\
\hline
\multirow{4}{*}{15} & \multirow{4}{*}{BERT (top\_k = 5)} 
    & Binary  & 97 & 40.5 & 139 & 88 & 56.5 \\
    & & Greedy  & 97 & 52.1 & 207 & 155 & 44.9 \\
    & & 3-Nary  & 97 & 42.4 & 132 & 85 & 54.6 \\
    & & 6-Nary  & 97 & 39.1 & 138 & 89 & 57.9 \\
\hline
\end{tabular}%
}
\caption{Performance results for BERT replacement strategy in Yelp}
\end{table}

\begin{table}[h!]
\resizebox{\textwidth}{!}{%
\renewcommand{\arraystretch}{1.5}
\begin{tabular}{|c|c|c|c|c|c|c|}
\hline
K (No. of words) & Algorithm & Original Accuracy (\%) & ASR (\%) & Avg Query & Avg Query (For Atk Success) & Attack Accuracy (\%) \\
\hline
ALL & Dyn-N & 97 & 98 & 379 & 378 & 2 \\
\hline
3 & Dyn-N & 97 & 11 & 44 & 33 & 86 \\
\hline
5 & Dyn-N & 97 & 18 & 62 & 43 & 79 \\
\hline
15 & Dyn-N & 97 & 42 & 133 & 86 & 55 \\
\hline
\end{tabular}%
}
\caption{Performance results for Dynamic-N in Yelp}
\end{table}

\begin{table}[h!]
\resizebox{\textwidth}{!}{%
\renewcommand{\arraystretch}{1.5}
\begin{tabular}{|c|c|c|c|c|c|c|c|}
\hline
K (No. of words) & Algorithm & Original Accuracy (\%) & ASR (\%) & Avg Query & Avg Query (For Atk Success) & Avg Query for fail & Attack Accuracy (\%) \\
\hline
ALL & Binary & 97 & 98 & 366 & 368 & 305 & 2 \\
& 3-Nary & 97 & 98 & 347 & 345 & 414 & 2 \\
& 6-Nary & 97 & 98 & 335 & 332 & 475 & 2 \\
\hline
3 & Binary & 97 & 11 & 47 & 33 & 49 & 86 \\
& 3-Nary & 97 & 12 & 45 & 32 & 47 & 85 \\
& 6-Nary & 97 & 12 & 47 & 33 & 49 & 86 \\
\hline
5 & Binary & 97 & 17 & 64 & 43 & 69 & 80 \\
& 3-Nary & 97 & 17 & 62 & 42 & 66 & 80 \\
& 6-Nary & 97 & 17 & 63 & 42 & 67 & 80 \\
\hline
15 & Binary & 97 & 40 & 130 & 85 & 161 & 57 \\
& 3-Nary & 97 & 39 & 127 & 81 & 158 & 58 \\
& 6-Nary & 97 & 42 & 126 & 83 & 158 & 56 \\
\hline
\end{tabular}%
}
\caption{Performance results for N-Nary in Hybrid(5\%) in Yelp}
\end{table}

\begin{table}[h!]
\resizebox{\textwidth}{!}{%
\renewcommand{\arraystretch}{1.5}
\begin{tabular}{|c|c|c|c|c|c|c|c|}
\hline
K (No. of words) & Algorithm & Original Accuracy (\%) & ASR (\%) & Avg Query & Avg Query (For Atk Success) & Avg Query for fail & Attack Accuracy (\%) \\
\hline
ALL & Binary & 97 & 97 & 364 & 363 & 410 & 2 \\
& 3-Nary & 97 & 97 & 347 & 346 & 404 & 2 \\
& 6-Nary & 97 & 97 & 334 & 331 & 475 & 2 \\
\hline
3 & Binary & 97 & 11 & 50 & 34 & 53 & 86 \\
& 3-Nary & 97 & 11 & 46 & 33 & 47 & 86 \\
& 6-Nary & 97 & 12 & 47 & 34 & 48 & 85 \\
\hline
5 & Binary & 97 & 17 & 68 & 41 & 73 & 81 \\
& 3-Nary & 97 & 17 & 62 & 42 & 66 & 80 \\
& 6-Nary & 97 & 17 & 63 & 42 & 67 & 80 \\
\hline
15 & Binary & 97 & 38 & 131 & 83 & 162 & 59 \\
& 3-Nary & 97 & 39 & 125 & 79 & 157 & 58 \\
& 6-Nary & 97 & 42 & 125 & 83 & 157 & 55 \\
\hline
\end{tabular}%
}
\caption{Performance results for N-Nary in Hybrid(10\%) in Yelp}
\end{table}

\begin{table}[h!]
\resizebox{\textwidth}{!}{%
\renewcommand{\arraystretch}{1.5}
\begin{tabular}{|c|c|c|c|c|c|c|c|}
\hline
K (No. of words) & Algorithm & Original Accuracy (\%) & ASR (\%) & Avg Query & Avg Query (For Atk Success) & Avg Query for fail & Attack Accuracy (\%) \\
\hline
ALL & Binary & 97 & 95 & 366 & 365 & 413 & 2 \\
& 3-Nary & 97 & 95 & 348 & 349 & 268 & 2 \\
& 6-Nary & 97 & 95 & 336 & 333 & 472 & 2 \\
\hline
3 & Binary & 97 & 12 & 61 & 40 & 64 & 85 \\
& 3-Nary & 97 & 11 & 58 & 37 & 61 & 86 \\
& 6-Nary & 97 & 12 & 70 & 40 & 74 & 85 \\
\hline
5 & Binary & 97 & 17 & 79 & 44 & 87 & 80 \\
& 3-Nary & 97 & 18 & 75 & 45 & 81 & 80 \\
& 6-Nary & 97 & 18 & 88 & 52 & 96 & 79 \\
\hline
15 & Binary & 97 & 37 & 139 & 81 & 174 & 60 \\
& 3-Nary & 97 & 37 & 136 & 81 & 170 & 60 \\
& 6-Nary & 97 & 39 & 147 & 87 & 187 & 58 \\
\hline
\end{tabular}%
}
\caption{Performance results for N-Nary in Hybrid(20\%) in Yelp}
\end{table}

\begin{table}[h!]
\resizebox{\textwidth}{!}{%
\renewcommand{\arraystretch}{1.5}
\begin{tabular}{|c|c|c|c|c|c|c|c|}
\hline
K (No. of words) & Algorithm & Original Accuracy (\%) & ASR (\%) & Avg Query & Avg Query (For Atk Success) & Avg Query for fail & Attack Accuracy (\%) \\
\hline
ALL & Binary & 97 & 95 & 368 & 366 & 465 & 2 \\
& 3-Nary & 97 & 95 & 347 & 349 & 267 & 2 \\
& 6-Nary & 97 & 95 & 336 & 333 & 427 & 2 \\
\hline
3 & Binary & 97 & 12 & 82 & 46 & 87 & 85 \\
& 3-Nary & 97 & 11 & 58 & 37 & 61 & 86 \\
& 6-Nary & 97 & 12 & 70 & 40 & 74 & 85 \\
\hline
5 & Binary & 97 & 18 & 98 & 53 & 109 & 79 \\
& 3-Nary & 97 & 18 & 75 & 45 & 81 & 80 \\
& 6-Nary & 97 & 18 & 88 & 52 & 96 & 79 \\
\hline
15 & Binary & 97 & 39 & 154 & 92 & 197 & 58 \\
& 3-Nary & 97 & 37 & 136 & 81 & 170 & 60 \\
& 6-Nary & 97 & 39 & 147 & 87 & 187 & 58 \\
\hline
\end{tabular}%
}
\caption{Performance results for N-Nary in Hybrid(30\%) in Yelp}
\end{table}

\begin{table}[h!]
\resizebox{\textwidth}{!}{%
\renewcommand{\arraystretch}{1.5}
\begin{tabular}{|c|c|c|c|c|c|c|c|}
\hline
K (No. of words) & Algorithm & Original Accuracy (\%) & ASR (\%) & Avg Query & Avg Query (For Atk Success) & Avg Query for fail & Attack Accuracy (\%) \\
\hline
ALL & Binary & 97 & 95 & 367 & 366 & 418 & 2 \\
& 3-Nary & 97 & 95 & 356 & 354 & 400 & 2 \\
& 6-Nary & 97 & 95 & 337 & 335 & 441 & 2 \\
\hline
3 & Binary & 97 & 12 & 82 & 46 & 87 & 85 \\
& 3-Nary & 97 & 12 & 95 & 53 & 101 & 85 \\
& 6-Nary & 97 & 13 & 70 & 41 & 74 & 84 \\
\hline
5 & Binary & 97 & 18 & 98 & 53 & 109 & 79 \\
& 3-Nary & 97 & 19 & 110 & 62 & 122 & 78 \\
& 6-Nary & 97 & 18 & 88 & 51 & 96 & 79 \\
\hline
15 & Binary & 97 & 39 & 154 & 92 & 197 & 58 \\
& 3-Nary & 97 & 41 & 165 & 102 & 213 & 56 \\
& 6-Nary & 97 & 39 & 147 & 87 & 187 & 58 \\
\hline
\end{tabular}%
}
\caption{Performance results for N-Nary in Hybrid(40\%) in Yelp}
\end{table}

\begin{table}[h!]
\resizebox{\textwidth}{!}{%
\renewcommand{\arraystretch}{1.5}
\begin{tabular}{|c|c|c|c|c|c|c|}
\hline
K (No. of words) & Algorithm & Original Accuracy (\%) & ASR (\%) & Avg Query & Avg Query (For Atk Success) & Attack Accuracy (\%) \\
\hline
ALL & DynN+Hyb & 97 & 95 & 321 & 320 & 2 \\
\hline
3 & DynN+Hyb & 97 & 12 & 22 & 17 & 86 \\
\hline
5 & DynN+Hyb & 97 & 18 & 35 & 25 & 79 \\
\hline
15 & DynN+Hyb & 97 & 42 & 95 & 60 & 55 \\
\hline
\end{tabular}%
}
\caption{Performance results for Dynamic-N+Hybrid with Split Threshold 5\% in Yelp}
\end{table}

\begin{table}[h!]
\resizebox{\textwidth}{!}{%
\renewcommand{\arraystretch}{1.5}
\begin{tabular}{|c|c|c|c|c|c|c|}
\hline
K (No. of words) & Algorithm & Original Accuracy (\%) & ASR (\%) & Avg Query & Avg Query (For Atk Success) & Attack Accuracy (\%) \\
\hline
ALL & DynN+Hyb & 97 & 95 & 321 & 319 & 2 \\
\hline
3 & DynN+Hyb & 97 & 12 & 22 & 18 & 86 \\
\hline
5 & DynN+Hyb & 97 & 18 & 34 & 25 & 79 \\
\hline
15 & DynN+Hyb & 97 & 42 & 95 & 60 & 55 \\
\hline
\end{tabular}%
}
\caption{Performance results for Dynamic-N+Hybrid with Split Threshold 10\% in Yelp}
\end{table}

\begin{table}[h!]
\resizebox{\textwidth}{!}{%
\renewcommand{\arraystretch}{1.5}
\begin{tabular}{|c|c|c|c|c|c|c|}
\hline
K (No. of words) & Algorithm & Original Accuracy (\%) & ASR (\%) & Avg Query & Avg Query (For Atk Success) & Attack Accuracy (\%) \\
\hline
ALL & DynN+Hyb & 97 & 95 & 321 & 319 & 2 \\
\hline
3 & DynN+Hyb & 97 & 12 & 22 & 17 & 86 \\
\hline
5 & DynN+Hyb & 97 & 18 & 34 & 25 & 79 \\
\hline
15 & DynN+Hyb & 97 & 42 & 95 & 59 & 55 \\
\hline
\end{tabular}%
}
\caption{Performance results for Dynamic-N+Hybrid with Split Threshold 20\% in Yelp}
\end{table}

\begin{table}[h!]
\resizebox{\textwidth}{!}{%
\renewcommand{\arraystretch}{1.5}
\begin{tabular}{|c|c|c|c|c|c|c|}
\hline
K (No. of words) & Algorithm & Original Accuracy (\%) & ASR (\%) & Avg Query & Avg Query (For Atk Success) & Attack Accuracy (\%) \\
\hline
ALL & DynN+Hyb & 97 & 95 & 321 & 319 & 2 \\
\hline
3 & DynN+Hyb & 97 & 12 & 22 & 18 & 86 \\
\hline
5 & DynN+Hyb & 97 & 18 & 35 & 25 & 79 \\
\hline
15 & DynN+Hyb & 97 & 42 & 95 & 60 & 55 \\
\hline
\end{tabular}%
}
\caption{Performance results for Dynamic-N+Hybrid with Split Threshold 30\% in Yelp}
\end{table}

\begin{table}[h!]
\resizebox{\textwidth}{!}{%
\renewcommand{\arraystretch}{1.5}
\begin{tabular}{|c|c|c|c|c|c|c|}
\hline
K (No. of words) & Algorithm & Original Accuracy (\%) & ASR (\%) & Avg Query & Avg Query (For Atk Success) & Attack Accuracy (\%) \\
\hline
ALL & DynN+Hyb & 96.91 & 94.9 & 323 & 321 & 2 \\
\hline
3 & DynN+Hyb & 97 & 11.4 & 22 & 17 & 86 \\
\hline
5 & DynN+Hyb & 97 & 17.6 & 34 & 25 & 79 \\
\hline
15 & DynN+Hyb & 96.91 & 41.8 & 94 & 59 & 55 \\
\hline
\end{tabular}%
}
\caption{Performance results for Dynamic-N+Hybrid with Split Threshold 40\% in Yelp}
\end{table}

\begin{table}[h!]
\resizebox{\textwidth}{!}{%
\renewcommand{\arraystretch}{1.5}
\begin{tabular}{|c|c|c|c|c|c|c|}
\hline
K (No. of words) & Algorithm & Original Accuracy (\%) & ASR (\%) & Avg Query & Avg Query (For Atk Success) & Attack Accuracy (\%) \\
\hline
ALL & Binary & 97 & 94 & 328 & 327 & 3 \\
 & 3-Nary & 97 & 93.6 & 321 & 316 & 3.4 \\
 & 6-Nary & 97 & 94.2 & 337 & 332 & 2.8 \\
\hline
3 & Binary & 97 & 11.4 & 46 & 31 & 85.6 \\
 & 3-Nary & 97 & 11.3 & 45 & 29 & 85.7 \\
& 6-Nary & 97 & 11.3 & 47 & 34 & 85.7 \\
\hline
5 & Binary & 97 & 17.3 & 64 & 41 & 79.7 \\
& 3-Nary & 97 & 17.4 & 62 & 42 & 79.6 \\
 & 6-Nary & 97 & 17.1 & 65 & 44 & 79.9 \\
\hline
15 & Binary & 97 & 40 & 127 & 83 & 57 \\
 & 3-Nary & 97 & 40.4 & 126 & 81 & 56.6 \\
 & 6-Nary & 97 & 39.9 & 130 & 86 & 57.1 \\
\hline
\end{tabular}%
}
\caption{Performance results for Hybrid algorithm(5\%) at Sentence Level in Yelp}
\end{table}

\begin{table}[h!]
\resizebox{\textwidth}{!}{%
\renewcommand{\arraystretch}{1.5}
\begin{tabular}{|c|c|c|c|c|c|c|}
\hline
K (No. of words) & Algorithm & Original Accuracy (\%) & ASR (\%) & Avg Query & Avg Query (For Atk Success) & Attack Accuracy (\%) \\
\hline
ALL & Binary & 97 & 94 & 328 & 327 & 3 \\
& 3-Nary & 97 & 94.2 & 321 & 320 & 2.8 \\
& 6-Nary & 97 & 93.9 & 331 & 325 & 3.1 \\
\hline
3 & Binary & 97 & 11 & 47 & 30 & 86 \\
& 3-Nary & 97 & 11.7 & 46 & 30 & 85.3 \\
& 6-Nary & 97 & 11.4 & 46 & 34 & 85.6 \\
\hline
5 & Binary & 97 & 17.7 & 65 & 42 & 79.3 \\
 & 3-Nary & 97 & 17.9 & 62 & 43 & 79.1 \\
 & 6-Nary & 97 & 17.6 & 63 & 44 & 79.4 \\
\hline
15 & Binary & 97 & 39.3 & 127 & 82 & 57.7 \\
& 3-Nary & 97 & 40.1 & 124 & 78 & 56.9 \\
& 6-Nary & 97 & 39.7 & 127 & 84 & 57.3 \\
\hline
\end{tabular}%
}
\caption{Performance results for Hybrid algorithm(10\%) at Sentence Level in Yelp}
\end{table}

\newpage
\onecolumn

\section{AGNews Tables}

\begin{table}[h!]
\resizebox{\textwidth}{!}{%
\renewcommand{\arraystretch}{1.5}
\begin{tabular}{|c|c|c|c|c|c|c|}
\hline
K (No. of words) & Algorithm & Original Accuracy (\%) & ASR (\%) & Avg Query & Avg Query (For Atk Success) & Attack Accuracy (\%) \\
\hline
ALL & Binary & 95 & 73 & 185 & 154 & 25 \\
& Greedy & 95 & 73 & 185 & 154 & 25 \\
& 3-Nary & 95 & 72 & 202 & 162 & 26 \\
& 6-Nary & 95 & 76 & 304 & 253 & 23 \\
\hline
3 & Binary & 95 & 7 & 50 & 44 & 88 \\
& Greedy & 95 & 7 & 50 & 44 & 88 \\
& 3-Nary & 95 & 7 & 33 & 29 & 88 \\
& 6-Nary & 95 & 6 & 37 & 29 & 89 \\
\hline
5 & Binary & 95 & 12 & 61 & 53 & 83 \\
& Greedy & 95 & 12 & 61 & 53 & 83 \\
& 3-Nary & 95 & 11 & 49 & 34 & 85 \\
& 6-Nary & 95 & 10 & 53 & 39 & 85 \\
\hline
15 & Binary & 95 & 35 & 117 & 89 & 60 \\
& Greedy & 95 & 35 & 117 & 89 & 60 \\
& 3-Nary & 95 & 31 & 118 & 81 & 64 \\
& 6-Nary & 95 & 26 & 116 & 77 & 69 \\
\hline
\end{tabular}}
\caption{Performance results for N-Nary in AGNews}
\label{table:metrics}
\end{table}

\begin{table}[h!]
\resizebox{\textwidth}{!}{%
\renewcommand{\arraystretch}{1.5}
\begin{tabular}{|c|c|c|c|c|c|c|}
\hline
K (No. of words) & Algorithm & Original Accuracy (\%) & ASR (\%) & Avg Query & Avg Query (For Atk Success) & Attack Accuracy (\%) \\
\hline
ALL & Binary & 95 & 73 & 178 & 140 & 26  \\
& 3-Nary & 95 & 72 & 178 & 139 & 26  \\
& 6-Nary & 95 & 77 & 275 & 220 & 23 \\
\hline
3 & Binary & 95 & 6 & 18 & 16 & 89 \\
& 3-Nary & 95 & 7 & 18 & 16 & 88 \\
& 6-Nary & 95 & 6 & 16 & 12 & 89 \\
\hline
5 & Binary & 95 & 9 & 32 & 26 & 86 \\
& 3-Nary & 95 & 11 & 31 & 20 & 85 \\
& 6-Nary & 95 & 10 & 28 & 20 & 85 \\
\hline
15 & Binary & 95 & 32 & 96 & 63 & 63 \\
& 3-Nary & 95 & 33 & 96 & 62 & 64  \\
& 6-Nary & 95 & 28 & 89 & 56 & 68  \\
\hline
\end{tabular}}
\caption{Performance results for AGNews with Hybrid 5\%}
\label{table:metrics_with_threshold}
\end{table}

\begin{table}[h!]
\resizebox{\textwidth}{!}{%
\renewcommand{\arraystretch}{1.5}
\begin{tabular}{|c|c|c|c|c|c|c|}
\hline
K (No. of words) & Algorithm & Original Accuracy (\%) & ASR (\%) & Avg Query & Avg Query (For Atk Success) & Attack Accuracy (\%) \\
\hline
ALL & Binary & 95 & 72 & 178 & 140 & 27 \\
& 3-Nary & 95 & 72 & 178 & 138 & 27  \\
& 6-Nary & 95 & 76 & 277 & 221 & 23  \\
\hline
3 & Binary & 95 & 6 & 18 & 15 & 89  \\
& 3-Nary & 95 & 7 & 18 & 16 & 88  \\
& 6-Nary & 95 & 6 & 16 & 12 & 89  \\
\hline
5 & Binary & 95 & 9 & 32 & 25 & 86  \\
& 3-Nary & 95 & 11 & 31 & 21 & 85 \\
& 6-Nary & 95 & 11 & 28 & 20 & 85  \\
\hline
15 & Binary & 95 & 32 & 96 & 63 & 63  \\
& 3-Nary & 95 & 33 & 96 & 62 & 64 \\
& 6-Nary & 95 & 28 & 89 & 55 & 69  \\
\hline
\end{tabular}}
\caption{Performance Metrics for AGNews with Hybrid 10\%}
\label{table:metrics_with_threshold}
\end{table}

\begin{table}[h!]
\resizebox{\textwidth}{!}{%
\renewcommand{\arraystretch}{1.5}
\begin{tabular}{|c|c|c|c|c|c|c|}
\hline
K (No. of words) & Algorithm & Original Accuracy (\%) & ASR (\%) & Avg Query & Avg Query (For Atk Success) & Attack Accuracy (\%)  \\
\hline
ALL & Binary & 95 & 74 & 178 & 141 & 26 \\
& 3-Nary & 95 & 72 & 178 & 138 & 27 \\
& 6-Nary & 95 & 76 & 278 & 224 & 23  \\
\hline
3 & Binary & 95 & 6 & 18 & 16 & 89  \\
& 3-Nary & 95 & 7 & 18 & 16 & 88  \\
& 6-Nary & 95 & 6 & 16 & 13 & 89 \\
\hline
5 & Binary & 95 & 9 & 32 & 27 & 86  \\
& 3-Nary & 95 & 11 & 31 & 21 & 85 \\
& 6-Nary & 95 & 11 & 28 & 21 & 85  \\
\hline
15 & Binary & 95 & 33 & 97 & 64 & 63  \\
& 3-Nary & 95 & 33 & 96 & 63 & 64  \\
& 6-Nary & 95 & 28 & 89 & 56 & 69 \\
\hline
\end{tabular}}
\caption{Performance results for AGNews with Hybrid 20\%}
\label{table:metrics_with_threshold}
\end{table}

\begin{table}[h!]
\resizebox{\textwidth}{!}{%
\renewcommand{\arraystretch}{1.5}
\begin{tabular}{|c|c|c|c|c|c|c|c|}
\hline
K (No. of words) & Algorithm & Original Accuracy (\%) & ASR (\%) & Avg Query & Avg Query (For Atk Success) & Attack Accuracy (\%) \\
\hline
ALL & Binary & 95 & 74 & 179 & 141 & 27 \\
& 3-Nary & 95 & 72 & 179 & 139 & 27\\
& 6-Nary & 95 & 76 & 277 & 220 & 23 \\
\hline
3 & Binary & 95 & 6 & 18 & 16 & 89  \\
& 3-Nary & 95 & 7 & 18 & 16 & 88  \\
& 6-Nary & 95 & 6 & 16 & 12 & 89  \\
\hline
5 & Binary & 95 & 9 & 32 & 26 & 86  \\
& 3-Nary & 95 & 11 & 31 & 21 & 85  \\
& 6-Nary & 95 & 10 & 28 & 20 & 85  \\
\hline
15 & Binary & 95 & 33 & 97 & 64 & 63  \\
& 3-Nary & 95 & 33 & 96 & 63 & 64  \\
& 6-Nary & 95 & 28 & 89 & 56 & 69  \\
\hline
\end{tabular}}
\caption{Performance results for AGNews with Hybrid 30\%}
\label{table:metrics_with_threshold}
\end{table}

\begin{table}[h!]
\resizebox{\textwidth}{!}{%
\renewcommand{\arraystretch}{1.5}
\begin{tabular}{|c|c|c|c|c|c|c|}
\hline
K (No. of words) & Algorithm & Original Accuracy (\%) & ASR (\%) & Avg Query & Avg Query (For Atk Success) & Attack Accuracy (\%)\\
\hline
ALL & Binary & 95 & 74 & 179 & 141 & 27  \\
& 3-Nary & 95 & 73 & 178 & 140 & 26 \\
& 6-Nary & 95 & 76 & 277 & 218 & 23  \\
\hline
3 & Binary & 95 & 6 & 18 & 15 & 89  \\
& 3-Nary & 95 & 7 & 18 & 17 & 88  \\
& 6-Nary & 95 & 6 & 16 & 12 & 89  \\
\hline
5 & Binary & 95 & 9 & 32 & 26 & 86 \\
& 3-Nary & 95 & 11 & 31 & 22 & 85  \\
& 6-Nary & 95 & 10 & 28 & 20 & 85  \\
\hline
15 & Binary & 95 & 33 & 97 & 64 & 64  \\
& 3-Nary & 95 & 33 & 96 & 63 & 64  \\
& 6-Nary & 95 & 28 & 89 & 56 & 69  \\
\hline
\end{tabular}
}
\caption{Performance results for AGNews with Hybrid 40\%}
\label{table:metrics_with_threshold_40}
\end{table}

\begin{table}[h!]
\resizebox{\textwidth}{!}{%
\renewcommand{\arraystretch}{1.5}
\begin{tabular}{|c|c|c|c|c|c|c|}
\hline
K (No. of words) & Algorithm & Original Accuracy (\%) & ASR (\%) & Avg Query & Avg Query (For Atk Success) & Attack Accuracy (\%) \\
\hline
ALL & Dyn N & 95 & 74 & 209 & 169 & 27 \\
\hline
3 & Dyn N & 95 & 7 & 33 & 28 & 88 \\
\hline
5 & Dyn N & 95 & 11 & 49 & 35 & 85 \\
\hline
15 & Dyn N & 95 & 26 & 116 & 77 & 69 \\
\hline
\end{tabular}%
}
\caption{Performance results for AGNews with Dynamic-N}
\end{table}

\begin{table}[ht]
\centering
\resizebox{\textwidth}{!}{%
\renewcommand{\arraystretch}{1.5}
\begin{tabular}{|c|c|c|c|c|c|c|}
\hline
K (No. of words) & Algorithm & Original Accuracy (\%) & ASR (\%) & Avg Query & Avg Query (For Atk Success) & Attack Accuracy (\%) \\
\hline
ALL & Dyn N + Hyb & 95 & 72 & 179 & 141 & 27 \\
\hline
3 & Dyn N + Hyb & 95 & 7 & 18 & 16 & 88 \\
\hline
5 & Dyn N + Hyb & 95 & 11 & 31 & 21 & 85 \\
\hline
15 & Dyn N + Hyb & 95 & 26 & 89 & 55 & 69 \\
\hline
\end{tabular}%
}
\caption{Performance results for AGNews with Dynamic N + Hybrid(5\%) algorithm}
\end{table}

\begin{table}[h!]
\resizebox{\textwidth}{!}{%
\renewcommand{\arraystretch}{1.5}
\begin{tabular}{|c|c|c|c|c|c|c|}
\hline
K (No. of words) & Algorithm & Original Accuracy (\%) & ASR (\%) & Avg Query & Avg Query (For Atk Success) & Attack Accuracy (\%) \\
\hline
ALL & Dyn N + Hyb & 95 & 73 & 179 & 141 & 27 \\
\hline
3 & Dyn N + Hyb & 95 & 7 & 18 & 16 & 88 \\
\hline
5 & Dyn N + Hyb & 95 & 10 & 31 & 21 & 85 \\
\hline
15 & Dyn N + Hyb & 95 & 26 & 89 & 56 & 69 \\
\hline
\end{tabular}%
}
\caption{Performance results for AGNews with Dynamic N + Hybrid(10\%) algorithm}
\end{table}

\begin{table}[h!]
\resizebox{\textwidth}{!}{%
\renewcommand{\arraystretch}{1.5}
\begin{tabular}{|c|c|c|c|c|c|c|}
\hline
K (No. of words) & Algorithm & Original Accuracy (\%) & ASR (\%) & Avg Query & Avg Query (For Atk Success) & Attack Accuracy (\%) \\
\hline
ALL & DynN+Hyb & 95 & 73 & 179 & 140 & 27 \\
\hline
3 & DynN+Hyb & 95 & 7 & 18 & 17 & 88 \\
\hline
5 & DynN+Hyb & 95 & 10 & 31 & 22 & 85 \\
\hline
15 & DynN+Hyb & 95 & 26 & 89 & 55 & 69 \\
\hline
\end{tabular}%
}
\caption{Performance results for AGNews with Dynamic N + Hybrid(20\%) algorithm}
\end{table}

\begin{table}[ht]
\resizebox{\textwidth}{!}{%
\renewcommand{\arraystretch}{1.5}
\begin{tabular}{|c|c|c|c|c|c|c|}
\hline
K (No. of words) & Algorithm & Original Accuracy (\%) & ASR (\%) & Avg Query & Avg Query (For Atk Success) & Attack Accuracy (\%) \\
\hline
ALL & DynN+Hyb & 95 & 72 & 178 & 141 & 26 \\
\hline
3 & DynN+Hyb & 95 & 7 & 18 & 16 & 88 \\
\hline
5 & DynN+Hyb & 95 & 10 & 31 & 20 & 85 \\
\hline
15 & DynN+Hyb & 95 & 26 & 89 & 55 & 69 \\
\hline
\end{tabular}%
}
\caption{Performance results for AGNews with Dynamic N + Hybrid(30\%) algorithm}
\end{table}

\begin{table}[ht]
\resizebox{\textwidth}{!}{%
\renewcommand{\arraystretch}{1.5}
\begin{tabular}{|c|c|c|c|c|c|c|}
\hline
K (No. of words) & Algorithm & Original Accuracy (\%) & ASR (\%) & Avg Query & Avg Query (For Atk Success) & Attack Accuracy (\%) \\
\hline
ALL & DynN+Hyb & 95 & 72 & 178 & 140 & 27 \\
\hline
3 & DynN+Hyb & 95 & 7 & 18 & 16 & 88 \\
\hline
5 & DynN+Hyb & 95 & 10 & 31 & 21 & 85 \\
\hline
15 & DynN+Hyb & 95 & 26 & 89 & 55 & 69 \\
\hline
\end{tabular}%
}
\caption{Performance results for AGNews with Dynamic N + Hybrid(40\%) algorithm}
\end{table}

\begin{table}[ht]
\resizebox{\textwidth}{!}{%
\renewcommand{\arraystretch}{1.5}
\begin{tabular}{|c|c|c|c|c|c|c|c|}
\hline
K (No. of words) & Algorithm & Original Accuracy (\%) & ASR (\%) & Avg Query & Avg Query (For Atk Success) & Avg Query for fail & Attack Accuracy (\%) \\
\hline
ALL & Binary & 95 & 72 & 200 & 160 & 304 & 26 \\
& 3-Nary & 95 & 70 & 201 & 164 & 303 & 25 \\
& 6-Nary & 95 & 73 & 218 & 172 & 338 & 26 \\
\hline
3 & Binary & 95 & 6 & 34 & 25 & 34 & 89 \\
& 3-Nary & 95 & 7 & 34 & 29 & 34 & 88 \\
& 6-Nary & 95 & 6 & 35 & 27 & 35 & 89 \\
\hline
5 & Binary & 95 & 9 & 50 & 37 & 51 & 86 \\
& 3-Nary & 95 & 11 & 49 & 36 & 50 & 84 \\
& 6-Nary & 95 & 10 & 52 & 38 & 53 & 85 \\
\hline
15 & Binary & 95 & 33 & 117 & 84 & 135 & 62 \\
& 3-Nary & 95 & 30 & 117 & 80 & 134 & 65 \\
& 6-Nary & 95 & 30 & 119 & 81 & 137 & 65 \\
\hline
\end{tabular}%
}
\caption{Performance results for AGNews with Sentence level + Hybrid(10\%)}
\end{table}

\begin{table}[h!]
\resizebox{\textwidth}{!}{%
\renewcommand{\arraystretch}{1.5}
\begin{tabular}{|c|c|c|c|c|c|c|c|}
\hline
K (No. of words) & Algorithm & Original Accuracy (\%) & ASR (\%) & Avg Query & Avg Query (For Atk Success) & Avg Query for fail & Attack Accuracy (\%) \\
\hline
ALL & Binary & 95 & 73 & 191 & 153 & 292 & 26 \\
& 3-Nary & 95 & 73 & 191 & 153 & 291 & 26 \\
& 6-Nary & 95 & 73 & 189 & 154 & 285 & 26 \\
\hline
3 & Binary & 95 & 6 & 34 & 24 & 35 & 89 \\
& 3-Nary & 95 & 7 & 34 & 27 & 34 & 88 \\
& 6-Nary & 95 & 6 & 33 & 25 & 34 & 89 \\
\hline
5 & Binary & 95 & 10 & 49 & 34 & 50 & 85 \\
& 3-Nary & 95 & 10 & 48 & 35 & 49 & 85 \\
& 6-Nary & 95 & 10 & 49 & 34 & 51 & 85 \\
\hline
15 & Binary & 95 & 32 & 112 & 80 & 128 & 63 \\
& 3-Nary & 95 & 32 & 111 & 80 & 126 & 63 \\
& 6-Nary & 95 & 32 & 112 & 79 & 130 & 63 \\
\hline
\end{tabular}%
}
\caption{Performance results for AGNews with Sentence level + Hybrid(20\%)}
\end{table}

\end{document}